\documentclass[hidelinks,12pt]{report}

\date{}
\usepackage{etoolbox}
\makeatletter
\patchcmd{\l@chapter}{1.0em}{0.8em}{}{}
\makeatother

\usepackage{amsmath} 
\usepackage{amssymb}
\usepackage{amsthm}
\usepackage[font={footnotesize,it}]{caption}
\usepackage{subcaption} 

\newtheorem{innercustomgeneric}{\customgenericname}
\providecommand{\customgenericname}{}
\newcommand{\newcustomtheorem}[2]{%
  \newenvironment{#1}[1]
  {%
   \renewcommand\customgenericname{#2}%
   \renewcommand\theinnercustomgeneric{##1}%
   \innercustomgeneric
  }
  {\endinnercustomgeneric}
}

\newcustomtheorem{customthm}{Theorem}
\newcustomtheorem{customlemma}{Lemma}

\usepackage{csquotes} 
\usepackage{algorithm} 
\usepackage{algpseudocode} 
\algnewcommand{\Initialize}[1]{%
  \State \textbf{Initialize:}
  \Statex \hspace*{\algorithmicindent}\parbox[t]{.8\linewidth}{\raggedright #1}
}
\usepackage[nottoc,notlot,notlof]{tocbibind} 
\algrenewcommand\algorithmicrequire{\textbf{Input:}}
\algrenewcommand\algorithmicensure{\textbf{Output:}}

\newcommand{\twonorm}[3]{\lVert #1#2#3 \rVert} 
\usepackage[italicdiff]{physics}  
\usepackage{enumitem}
\usepackage{indentfirst}
\usepackage{mathtools}
\usepackage{cite}

\usepackage{hyperref}
\usepackage{graphicx}
\graphicspath{{Images/}}
\usepackage{type1cm}
\usepackage[thinc]{esdiff}
\usepackage{amsfonts}
\usepackage{mathrsfs}
\usepackage{multirow} 
\usepackage{setspace}
\usepackage{xcolor}
\usepackage{booktabs} 
\usepackage{hhline} 
\usepackage{wallpaper} 
\CenterWallPaper{.18}{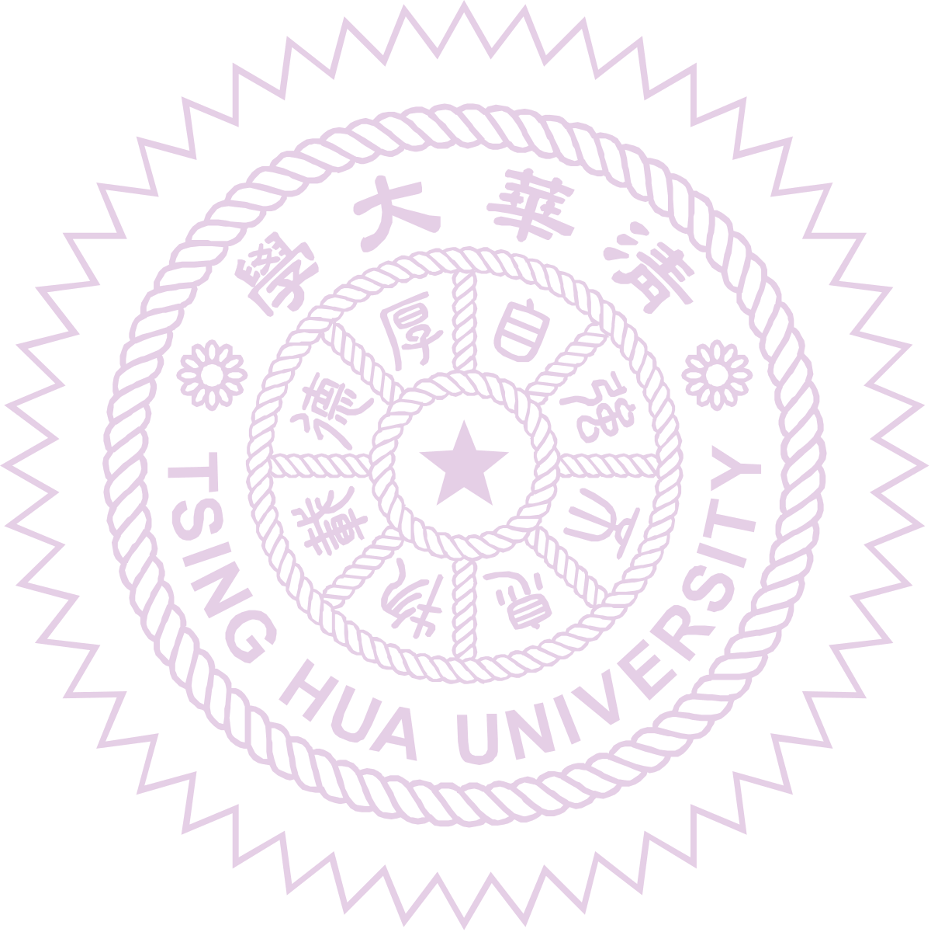}
\usepackage{tocloft}

\usepackage[top=3cm,bottom=1cm]{geometry}
\newcommand{\llapph}[1]{\phantom{00.00\%}\llap{#1}}

\def\R{{\cal R}}

\def\M{M}

\def\tr{{tr}}  
\def\tt{{tt}}

\def\N{{\cal N}}

\def\R{{\hbox{\mit I\kern-.2em R}}}
\def\N{{\hbox{\mit I\kern-.2em N}}}

\usepackage{lmodern}
\allowdisplaybreaks 
\usepackage{physics} 

\begin{document}

\begin{titlepage}
\vspace*{0.25cm}
\begin{center}
\Large \textbf{Institute of Statistics}\\[3pt] \textbf{College of Science}\\[3pt] \textbf{National Tsing Hua University}  \\[5pt] 
\LARGE \textbf{Master Thesis}
\end{center}

\vspace{100pt}

\begin{center}
{\LARGE
\mbox{\textbf{Gradient-based Quadratic Multiform Separation}}}
\end{center}

\vspace{100pt}
\begin{center}
{\fontsize{16}{18}\selectfont
\begin{tabular}{lll}
\textbf{Candidate ID :} & \textbf{108024507}&\\[5pt]
\textbf{Candidate :} & \textbf{Wen-Teng Chang}\\[5pt]
\textbf{Advisor :} & \textbf{Dr.\;Ching-Kang Ing}\\[5pt]
\textbf{Advisor :} & \textbf{Dr.\;Michael Fan}
\end{tabular}}
\end{center}

\vspace{60pt}

\begin{center}
\Large
\mbox{\textbf{July, 2021}}
\end{center}
\end{titlepage}

\pagenumbering{roman}

\begin{center}
\section*{Abstract}
\end{center}

Classification as a supervised learning concept is an important content in machine learning. It aims at categorizing a set of data into classes. There are several commonly-used classification methods nowadays such as k-nearest neighbors, random forest, and support vector machine. Each of them has its own pros and cons, and none of them is invincible for all kinds of problems. In this thesis, we focus on Quadratic Multiform Separation (QMS), a classification method recently proposed by Michael Fan et al. (2019). Its fresh concept, rich mathematical structure, and innovative definition of loss function set it apart from the existing classification methods. Inspired by QMS, we propose utilizing a gradient-based optimization method, Adam, to obtain a classifier that minimizes the QMS-specific loss function. In addition, we provide suggestions regarding model tuning through explorations of the relationships between hyperparameters and accuracies. Our empirical result shows that QMS performs as good as most classification methods in terms of accuracy. Its superior performance is almost comparable to those of gradient boosting algorithms that win massive machine learning competitions.

\vspace*{1cm}
\noindent Key Words: Quadratic Multiform Separation, QMS, Adam, Supervised Learning, Classification

\clearpage

\begin{center}
\section*{Acknowledgement}
\end{center}

Time flies, here comes to an end of my student career from bachelor to master at National Tsing Hua University. I appreciate all the learnings and happenings the campus provides, it nurtures me to become a better person in aspect of both knowledge and character.

First of all, I would like to say thank you to my parents. During my studies in Hsinchu which is away from my home, they barely show their worries and fully trust me of what I am doing and planning for the  future. Thank you, my mom and dad, for always being there backing me up during every ups and downs. 

Additionally, I am wholeheartedly grateful to have Prof.\;Ching-Kang Ing and Prof.\;Michael Fan as my thesis advisors. During these 2 years, Professor Ing always gives me his greatest support and encouragement on letting me to make my own personal career decisions including the participation of internship and overseas exchange. Furthermore, the thesis discussions with Professor Ing really inspires me a lot to think out of the box; Professor Fan always shows his patience on walking me through his invention, QMS, and the relevant technical knowledge, this kind of step-by-step coaching is what I really appreciate a lot. Without their careful guidances and illuminating instructions, this thesis couldn't reach its present form.

Last but not least, I am glad to be in part of NTHU STAT family to meet all the warm and intelligent friends. The moments that we share, discuss, and collaborate with each other are the time I cherish the most and never want to let go. I love such an environment and I will keep all these wonderful memories carefully in my mind.

\hfill Sincerely,

\hfill Wen-Teng Chang 

\hfill 2021/7/19

\clearpage
\tableofcontents

\clearpage

\listoffigures

\clearpage

\listoftables

\clearpage

\pagenumbering{arabic}

\chapter{Introduction}

\noindent In 1959, Arthur Samuel defined machine learning as a ``Field of study that gives computers the ability to learn without being explicitly programmed" \cite{1403109}.  In 1997, Tom M. Mitchell provided a widely quoted and more formal definition: ``A computer program is said to learn from experience E with respect to some class of tasks T and performance measure P, if its performance at tasks in T, as measured by P, improves with experience E" \cite{27308}. Nowadays, there are four main types of machine learning tasks which are Supervised learning, Unsupervised learning, Semi-supervised learning, and Reinforcement learning. 

\begin{itemize}
\item \textbf{Supervised learning}: The computer learns a general function aiming to map the input to output (labels of input) for a given dataset. The task of supervised learning can be divided into two segments which are classification and regression problems depending on the type of label being categorical or continuous.
\item \textbf{Unsupervised learning}: The main difference between supervised and unsupervised learning is whether the input is labelled or not. Therefore, the main goal of unsupervised learning is to discover hidden patterns in data or a means towards an end.
\item \textbf{Semi-supervised learning}: Semi-supervised learning falls between supervised and unsupervised learnings. It typically comprises a small amount of labelled data and a large amount of unlabelled data while training.
\item \textbf{Reinforcement learning}: Unlike supervised learning that there is a ``teacher" or complete dataset to work on, in the field of reinforcement learning, machine interacts with a dynamic environment and learns from its failure. 
\end{itemize}

In this thesis, we'll focus on classification problem in supervised learning. As we know that there are several classification methods existing nowadays including k-nearest neighbors (k-NN), random forest (RF), support vector machine (SVM), etc. Each of these methods has its own perspective for classification. For example, k-NN classifies a new data point based on its surroundings via majority vote since the author believes that the label of that data point can be described by its neighbors; SVM classifies a new data point according to the pre-constructed hyperplane due to the belief of a  high-dimensional projection of data can be perfectly splitted for different classes. Most of these commonly-used classification methods are maturely developed already. In 2019, Michael Fan et al. proposed a profoundly different method called Quadratic Multiform Separation (QMS) \cite{FanJanuary142021, FanApril142021}. QMS is fresh for its intuitive concept, abundant mathematical structure, and original loss function definition. It classifies a new data point into one of $m$ classes on the basis of pre-constructed $m$ buckets (member functions) because it is believed that a new data point can be described by its bucket name. A question arising naturally is that how to construct the buckets in QMS? We embed a gradient-based optimization method, Adam, to create the bucket for each class in the data by minimizing the specific loss function. These $m$ buckets together construct a classifier for conducting prediction. 

As other classification methods do, there are some hyperparameters that can be tuned in QMS for achieving better performance. Therefore, we'll further investigate their ($q$ and $\alpha$) properties and offer suggestions for model tuning. The empirical results show that gradient-based QMS performs as good as the mainstream classification methods in terms of accuracy. This success provides one with a new choice for the purpose of pursuing a high accuracy of prediction. Note that the datasets adopted for comparison are more of less balanced so that it is reasonable to consider accuracy as a suitable performance measurement.

The outline of the thesis is as follows. In Chapter 2, we introduce six popular machine learning algorithms that will be considered in future section for benchmarking. In Chapter 3, Quadratic Multiform Separation (QMS) is elaborated, this includes its core concept, mathematical structure, and the specific loss function. In Chapter 4, we incorporate a gradient-based optimization method, Adam, to find the classifier by minimizing the QMS-specific loss function. Furthermore, we also explore the relationships between hyperparameters and accuracies with the suggestions of model tuning provided. At last, in Chapter 5, QMS is implemented on five datasets and been compared with six outstanding machine learning algorithms to see how it stands out from the crowd. 

\chapter{Machine Learning Algorithms}
\noindent 
In this chapter, we briefly review six matured classification methods including k-nearest neighbors, logistic regression, random forest, support vector machine, extreme gradient boosting and artificial neural network. In addition, their performances will serve as comparisons to gradient-based QMS in Chapter 5. Suppose we have pairs $(x_{1}, y_{1}), (x_{2}, y_{2}), \dots, (x_{n}, y_{n})$, where $\{x_i\}_{i = 1,\ldots,n}$ denotes an observation (feature vector) and $\{y_i\}_{i=1,\ldots,n}$ denotes the label  among $m$ classes of the observation. Our goal is to classify a new data point $x_{new}$ to obtain its label $y_{new}$.
\section{k-Nearest Neighbors (k-NN)}

\noindent 
k-NN \cite{917808} is a data-driven algorithm which believes that a new sample can be classified based on the majority vote of nearest $k$ samples' labels, where $k$ is a user-defined hyperparameter and the term ``nearest" is in the aspect of distance. In general, $k$ should be set as an odd number to avoid the risk of a tie. 

\medskip \noindent
Learning Steps: 
		\begin{enumerate}
		\item Determine the number of nearest neighbors $k$ and the distance measurement such as Euclidean Distance, Manhattan distance, etc.
		\item Calculate all the distances between new sample and training data, then sort them to identify the closest $k$ points to $x_{new}$, represented by $N_0$.
		\item Obtain $$\hat{p}_{new,\,c} = Pr(y_{new} = c | X = x_{new}) = \frac{1}{k}\sum_{i\in N_0}I(y_i=c),\;\,\text{for}\;c = 1,\ldots,m$$ and classify $x_{new}$ as $$\hat{y}_{new} = \arg\max_{c=1,\ldots,m}\{\hat{p}_{new,\,c}\}$$.
		\end{enumerate}

\section{Logistic Regression}

\noindent 
Logistic regression \cite{917808} is similar to linear regression except for that it is used to solve a classification problem. There are two core concepts in logistic regression, one is logit and the other is sigmoid function which is shown below. 
	
	\begin{figure}[htbp]
   \centering
	\includegraphics[scale=0.4]{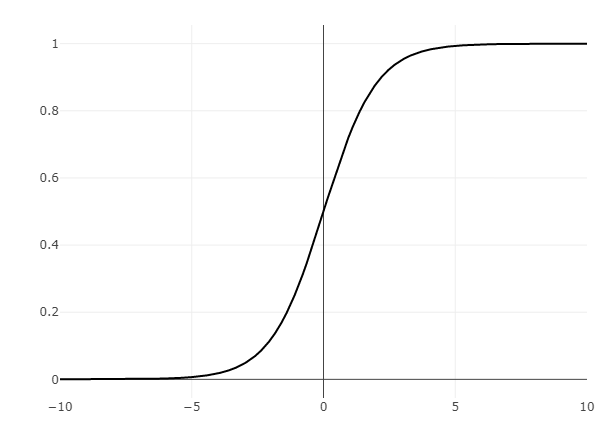}
	\caption{Sigmoid Function $(\frac{1}{1+e^{-x}})$}
	\label{fig:}
	\end{figure}

\noindent
The form of the logistic regression for the binary response variable $(y_i)$ goes like
$$\log \frac{P(y_i =1 | X = x_i)}{1-P(y_i =1 | X = x_i)} = x_i\beta,\;\, \text{for}\;  i = 1,\ldots,n$$
where $x_i$ is a feature vector, $\beta$ is the coefficient vector with the intercept, $\frac{P(y_i = 1 | X = x_i)}{1-P(y_i = 1 | X = x_i)}$ is odds ratio, and $\log(\frac{P(y_i =1 | X = x_i)}{1-P(y_i = 1 | X = x_i)})$ is named logit. After transpositioning the terms in above equation, we can obtain the following formula.
$$P(y_i=1 | X = x_i) = \frac{e^{x_i\beta}}{1+e^{x_i\beta}} =\frac{1}{1+e^{-x_i\beta}},\;\, \text{for}\;  i = 1,\ldots,n$$
Now, it is obvious that the sigmoid function mentioned at first being used to model the probability of $y_i = 1$. The coefficient vector $\beta$ is estimated by maximizing likelihood method given $y_i$ follows the bernoulli distribution. Moreover, we can transform the likelihood function by taking negative logarithm to obtain logloss (cross entropy). After calculating the partial derivative of logloss over $\beta$, the gradient descent takes into part to find the optimum of $\hat{\beta}$.

As for multi-class classification problem in logistic regression, we can first set a category as a ``pivot" and then the other $m-1$ outcomes are separately regressed against the pivot outcome. For example, if there are 3 categories $A, B, C$ and $A$ is set to be the pivot, we have
\begin{align*}
 \log \frac{P(y_i = B | X = x_i)}{P(y_i = A | X = x_i)} =  x_i\boldsymbol{\beta_B} &\Rightarrow P(y_i = B | X = x_i) = P(y_i = A | X = x_i) \times e^{x_i\boldsymbol{\beta_B}}\\
\log \frac{P(y_i = C | X = x_i)}{P(y_i = A | X = x_i)} =  x_i\boldsymbol{\beta_C} &\Rightarrow P(y_i = C | X = x_i) = P(y_i = A | X = x_i) \times e^{x_i\boldsymbol{\beta_C}}
\end{align*}
Using the fact that the sum of all 3 probabilities equals to one, we then obtain
$$P(y_i = A | X = x_i) = 1 - \sum_{k\in \{B,C\}} P(y_i = A| X = x_i) \times e^{x_i\boldsymbol{\beta_k}} \Rightarrow P(y_i = A | X = x_i) = \frac{1}{1+ \sum_{k\in \{B,C\}} e^{x_i\boldsymbol{\beta_k}}}$$ 
Therefore, we can find the rest of the probabilities as presented below.
\begin{align*}
P(y_i = B | X = x_i) &= \frac{e^{x_i\boldsymbol{\beta_B}}}{1+ \sum_{k\in \{B,C\}} e^{x_i\boldsymbol{\beta_k}}}\\
P(y_i = C | X = x_i) &= \frac{e^{x_i\boldsymbol{\beta_C}}}{1+ \sum_{k\in \{B,C\}} e^{x_i\boldsymbol{\beta_k}}}
\end{align*}

\medskip\noindent
Learning Steps:
\begin{enumerate}
\item Update the logloss by numerical method such as gradient descent to obtain the optimal estimated coefficients.
\item Obtain 
\begin{align*}
\hat{p}_{new,\,c^{'}} &= P(y_{new} = c^{'} | X = x_{new}) =\frac{1}{1+ \sum_{k} e^{x_{new}\boldsymbol{\beta_k}}},\;\text{where}\;\,c^{'}\;\,\text{is the pivot.} \\
\mathrlap{\hat{p}_{new,\,c}}\phantom{\hat{p}_{new,\,c^{'}}} &= P(y_{new} = c\;| X = x_{new}) =\frac{e^{x_{new}\boldsymbol{\beta_c}}}{1+ \sum_{k} e^{x_{new}\boldsymbol{\beta_k}}} ,\;\,\text{for}\;\,c = 1,\ldots,m\;\, \text{and}\;\, c \neq c^{'} 
\end{align*}
and classify $x_{new}$ as $$\hat{y}_{new} = \arg\max_{c=1,\ldots,m}\{\hat{p}_{new,\,c}\}$$
\end{enumerate}

\section{Random Forest (RF)}

\noindent 
As its name implies, random forest \cite{Breiman2001} is an ensemble learning method that operates by constructing a multitude of decision trees under random mechanism. A decision tree is drawn upside down with its root at the top. There are $3$ core elements in a decision tree which are internal node, edges, and external (leaf) node representing condition, branches, and decision, respectively. An external node is one without child branches, in contrast, internal node has at least one child branch. Below is a typical structure of a decision tree.
	
	\begin{figure}[htbp]
	\centering
	\includegraphics[scale=0.5]{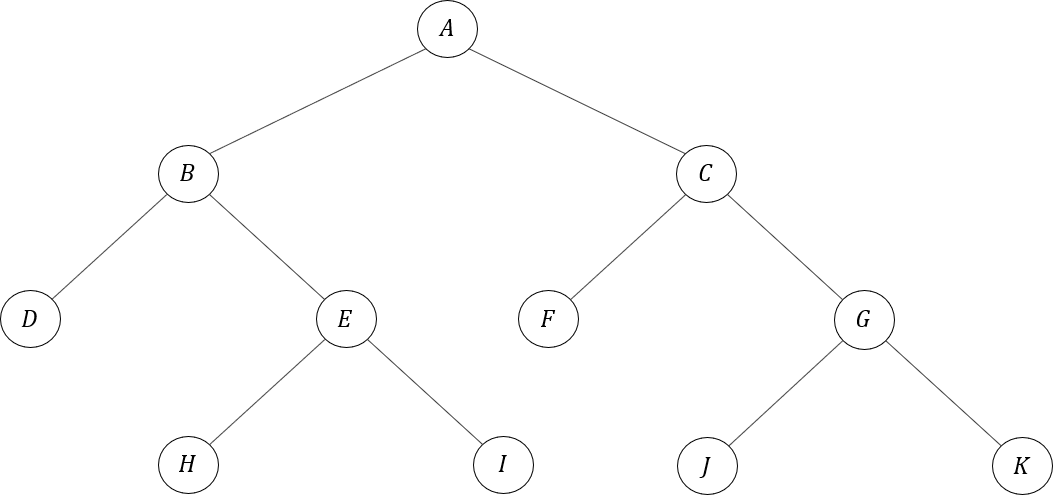}
	\caption{Decision Tree}
	\label{fig:figure2}
	\end{figure}
	
\noindent $A, B, C, E, G$ are internal nodes which are root node and $A, A, B, C$'s branches, respectively, and $D, H, I,  F, J, K$ are external nodes. Note that the number of external nodes is exactly one more than the number of internal nodes. 

As for random mechanism, there are two aspects of randomness, one is randomly sample (with replacement) a subset of training observations and the other is randomly select sufficient features, and these two together construct a single decision tree.  Once the feature pool and the training observations are specified, we then able to determine the best feature from the pool and the corresponding condition for each internal node. The term ``best" is in terms of maximizing the information gain which can be obtained by calculating the difference in gini impurity index or entropy between the node and its branches.

\medskip \noindent	
Learning Steps:
	\begin{enumerate}
	\item Select a criterion, gini index or entropy, to measure the goodness of partition  (feature + condition) in each internal nodes, where 
	$$gini\;index = 1 - \sum^m_{c=1}P_c^2 \in [0,\frac{m-1}{m}]$$ and
	$$entropy = -\sum^m_{c=1}P_c\log_2(P_c) \in [0,\log_2(m)]$$ where $P_c$ is the proportion of records after a partition that belong to class $c$. In practice, the chosen of the impurity criterion does not influence testing performance much \cite{Raileanu2004}.
	\item Tune the hyperparameters such as percentage of sample, percentage of features, maximum depth, etc, for constructing a tree with cross validation applied.
	\item Drop a new record all the way down to an external node, we can assign its class simply by taking a ``vote" of all the training data that belonged to that external node when the tree was grown \cite{2158511}. Since there are several trees in random forest, we have to repeat the process and assign the class for that observation by ``majority vote".
	\item  Obtain $$\hat{p}_{new,\,c} = \hat{f}_{rf,\;c}(X=x_{new}),\;\,\text{for}\;c = 1,\ldots,m$$  and classify $x_{new}$ as $$\hat{y}_{new} = \arg\max_{c=1,\ldots,m}\{\hat{p}_{new,\,c}\}$$
	\end{enumerate}
	
\section{Support Vector Machine (SVM)}

\begin{figure}[htbp]
	\centering
	\includegraphics[scale=0.5]{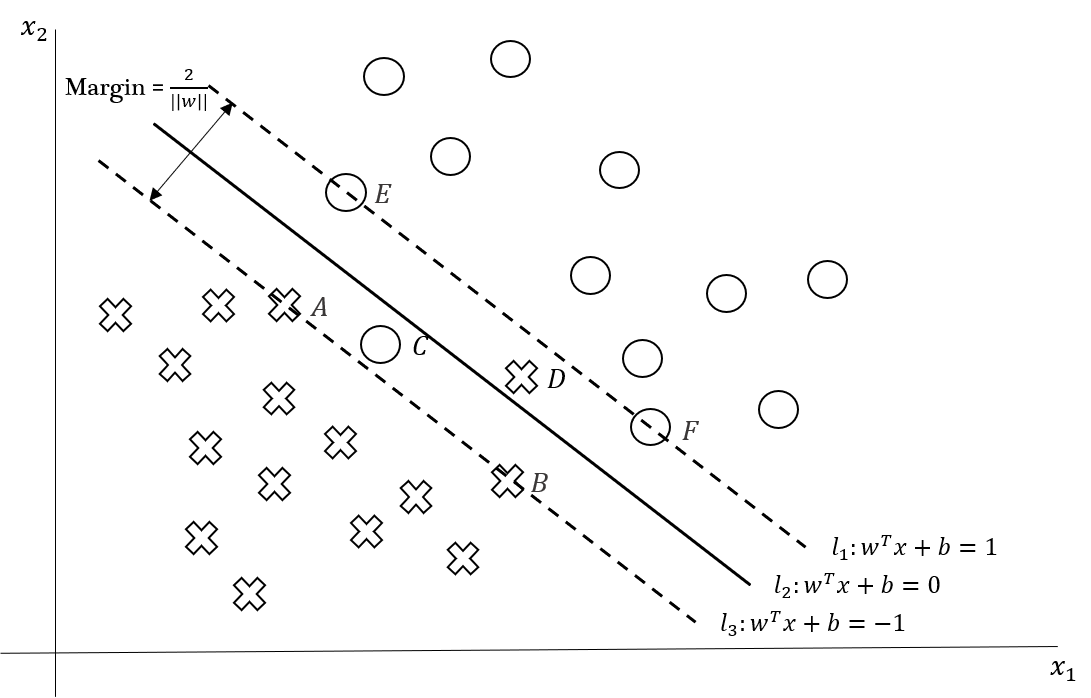}
	\caption{Support Vector Machine (2-class)}
	\label{fig:figure3}
	\end{figure}

\noindent 
Support vector machine \cite{Noble2006} is composed of 4 core elements, the separating hyperplane, the maximum-margin concept, the type of margin (hard or soft-margin) and the kernel function. The goal of SVM is to find a hyperplane that can maximize the margin between two auxiliary hyperplanes being $l_1$ and $l_3$ in Figure \ref{fig:figure3}, and the kernel function takes part in the situation that the original feature space is inseparable. Note that $l_2$ is the target hyperplane and it's simply a line in the 2-dimensional case. $A, B, E, F$ are called support vectors since they support the construction of $l_1$ and $l_3$. The difference between soft and hard-margin is that whether the tolerance term $\xi$ (so-called slack variable) is included in the optimization, this slack variable is a trade-off term between the violation of hyperplane and the size of margin utilized in soft margin SVM. That is to say, the inclusion of $\xi$ in soft-margin SVM is used to solve the potential overfitting problem derived from hard-margin one. Below is the optimization problem that is tackled by soft-margin SVM.
	$$\arg\min_{b,\;w,\;\xi}\frac{1}{2}w^Tw + C \sum_{i=1}^n\xi_i$$
	$$\text{subject to}\;\; y_i(\phi(x_i)^Tw+b) \geq 1-\xi_i$$ where $\xi_i\geq 0, \forall i=1,\ldots,n$, $C$ is a regularization hyperparameter and $\phi(\cdot)$ denotes the kernel trick.
	
As for multi-class classification problem in SVM, there are two methods,  one-vs-one (OVO) and one-vs-rest (OVR). OVO breaks down multi-class classification problem into  multiple binary classification cases $(C^m_2)$ and conduct majority vote. OVR divides the training dataset into a certain class $c$ and the rest, then by finding the $\text{SVM}_c$ that maximizes the corresponding function output (named confidence or probability), we can classify the new data as $c$. 
\clearpage
\noindent
Learning Steps:
	\begin{enumerate}
	\item Find a suitable kernel function to project the training data onto a high-dimensional space and penalty term $C$ to restrict the impact of $\xi$ in soft margin SVM. In practice, this is a trial and error process with cross-validation applied.
	\item Solve the above optimization problem and obtain the separating hyperplane. In fact, we usually transform the problem into its Lagrangian dual function.
	\item Classify $x_{new}$ as  $$\hat{y}_{new} = sign(w^T\phi(x_{new})+b)$$ for binary case.  As for multi-class classification problem, we can implement either OVO or OVR being mentioned above.
	\end{enumerate}
	
\section{eXtreme Gradient Boosting (XGBoost)}

\noindent 
XGBoost \cite{DBLP:journals/corr/ChenG16} is a boosting algorithm which is an ensemble method. It is a sequential one which combines several weak learners into a strong classifier. That is, it sequentially fits the residual from previous learner and produce a better one. The mathematical form of boosting concept is as follows.
$$
\hat{y_i}^{(t)} = \sum_{k=1}^t f_k(x_i) = \hat{y_i}^{(t-1)} + f_t(x_i)
$$
where $f_k$ is $k^{th}$ weak learner which can be either a tree or a linear model. 

XGBoost algorithm features its loss function $L^{(t)}$ which is written as
\begin{align*}
L^{(t)} &= \sum_{i=1}^n l(y_i,\hat{y_i}^{(t)}) + \sum_{k=1}^t\Omega(f_k)\\
&= \sum_{i=1}^n l(y_i,\hat{y_i}^{(t-1)} + f_t(x_i)) + \sum_{k=1}^t\Omega(f_k)\\
& = \sum_{i=1}^n l(y_i,\hat{y_i}^{(t-1)} + f_t(x_i)) + \Omega(f_t) + c
\end{align*}
where $l(y_i,\hat{y_i}^{(t)})$ is a differentiable convex function that can be mean square error, cross entropy, etc, and $c$ equals to $\sum_{k=1}^{t-1}\Omega(f_k)$ being a given constant when the training progress gets into the $t^{th}$ tree. Now, we can conduct Taylor expansion on $L^{(t)}$ to obtain
$$
L^{(t)} \approx \sum_{i=1}^n\left[g_if_t(x_i) + \frac{1}{2}h_if_t^2(x_i)\right] + \Omega(f_t)
$$
where $g_i = \pdv{l(y_i,\hat{y_i}^{(t-1)})}{\hat{y_i}^{(t-1)}}, h_i = \pdv[2]{l(y_i,\hat{y_i}^{(t-1)})}{(\hat{y_i}^{(t-1)})}$, in addition, the first term of Taylor expansion $l(y_i,\hat{y_i}^{(t-1)})$ and $c$ are omitted due to they have no effect on this optimization problem. Note that $g_i$ and $h_i$ are constants since $y_i$ and $\hat{y_i}^{(t-1)}$ are already known. 

The next question comes into our minds is what are the actual forms of predictor $f_t$ and the complexity function $\Omega(f_t)$. The author defines as below,
\begin{align*}
f_t(x) &= w_{q(x)}\\
\Omega(f_t) &= \gamma T + \frac{1}{2}\lambda\sum_{j=1}^T w_j^2
\end{align*}
where $w_{q(x)}\in \R^T, q:\R^p \rightarrow \{1,\ldots,T\}, T$ is number of leaves, $\gamma$ and $\lambda$ are regularization terms. 
		\begin{figure}[htbp]
		\centering
		\includegraphics[scale=0.5]{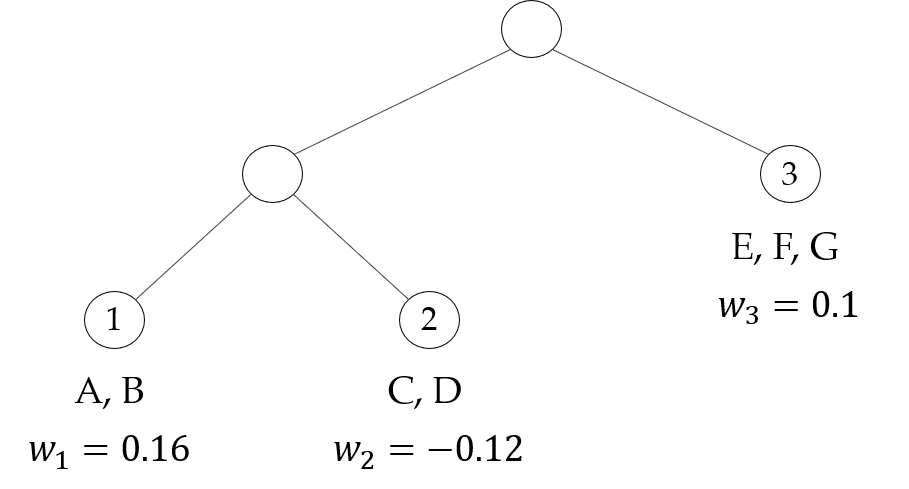}
		\caption{XGBoost}
		\label{fig:figure24}
		\end{figure}
		
\noindent For further elaboration, let's have a look on above Figure \ref{fig:figure24}, where $A,B,\ldots,G$ are observations. This is a tree with $T = 3$ and $q(A) = q(B) = 1, q(C)= q(D) = 2,$ etc. $w_{q(A)} = w_{q(B)} = 0.16$ serving as an addition of previous prediction of $A$ and $B$. Therefore, $\Omega = 3\gamma	+ \frac{1}{2}\lambda(0.0256+0.0144+0.01)$. 

Now, we can replace $f_t(x_i)$ and $\Omega(f_t)$ in $L^{(t)}$ with their explicit forms. Let's define a sample set $I_j = \{i\,|\,(q(x_i) = j\}$ which is a collection of $i$ when $q(x_i)$ equals to leaf $j$.
\begin{align*}
L^{(t)} &\approx \sum_{i=1}^n\left[g_if_t(x_i) + \frac{1}{2}h_if_t^2(x_i)\right] + \Omega(f_t)\\
 &= \sum_{i=1}^n\left[g_iw_{q(x_i)}+\frac{1}{2}h_iw_{q(x_i)}^2\right] + \gamma T + \frac{1}{2}\lambda\sum	_{j=1}^Tw_j^2\\
 &= \sum_{j=1}^T\left[(\sum_{i\in I_j}g_i)w_j + \frac{1}{2}(\sum_{i\in I_j}h_i + \lambda)w_j^2\right] + \gamma T
\end{align*}
\noindent For simplicity, let's denote $G_j = \sum_{i\in I_j}g_i$ and $H_j = \sum_{i\in I_j}h_i$. Since every leaf node is independent to each other, thus $L^{(t)}$ reaches its minimum only when each leaf node reaches its minimum. Furthermore, $w_j$ is the only parameter unknown, the optimal $w_j$ with respect to $L^{(t)}$ can be obtained with the following method.
\begin{align*}
(\sum_{i\in I_j}g_i)w_j + \frac{1}{2}(\sum_{i\in I_j}h_i + \lambda)w_j^2 
&= G_jw_j + \frac{1}{2}(H_j + \lambda)w_j^2 \\
&= \frac{1}{2}(H_j+\lambda)(w_j + \frac{G_j}{H_j+\lambda})^2 - \frac{1}{2}\frac{G_j^2}{H_j+\lambda} 
\end{align*}
\noindent Hence, when $w_j = w_j^* =  -\frac{G_j}{H_j+\lambda}$, the loss function is at its minimum with below form.
$$
L^{(t)} \approx -\frac{1}{2}\sum_{i=1}^T\frac{G_j^2}{H_j+\lambda}+\gamma T
$$ 
Now, we can define $Gain$ as a measurement for goodness of partition in XGBoost.
$$
Gain = \frac{1}{2}\left[\frac{G_L^2}{H_L+\lambda} + \frac{G_R^2}{H_R+\lambda} - \frac{(G_L+G_R)^2}{H_L+H_R+\lambda}\right] - \gamma
$$
\noindent where subscript $L$ and $R$ denote the left and right node, respectively. By applying Greedy Algorithm, we can find the best combination of feature and value for each nodes in terms of $Gain$ until the pre-determined tree's depth is met, then prune the nodes with negative gains.

To get insights of the $Gain$'s formula, we let $l(y_i, \hat{y_i}) = \frac{1}{2}\sum_{i\in I_j}(y_i - \hat{y_i})^2$ which is a loss function commonly-used in regression problem. We have $\dv{l(y_i, \hat{y_i})}{\hat{y_i}} = -\sum_{i\in I_j}(y_i - \hat{y_i})$ and $\dv[2]{l(y_i, \hat{y_i})}{\hat{y_i}} = n_{I_j}$. The $Gain$ with this specific loss function has the following form.
$$
Gain_{Reg} =  \frac{1}{2}\left[\frac{(-\sum_{i\in I_L}(y_i - \hat{y_i}))^2}{n_{I_L}+\lambda} + \frac{(-\sum_{i\in I_R}(y_i - \hat{y_i}))^2}{n_{I_R}+\lambda} - \frac{(-\sum_{i\in I_L}(y_i - \hat{y_i})-\sum_{i\in I_R}(y_i - \hat{y_i}))^2}{n_{I_L}+n_{I_R}+\lambda}\right] - \gamma
$$
The corresponding optimal output $w_j^*$ equals to $\frac{\sum_{i\in I_j}(y_i - \hat{y_i})}{n_{I_j}+\lambda}$.

\medskip\noindent
Learning Steps:
\begin{enumerate}
\item Find the best combinations of hyperparameters utilized in XGBoost algorithm including max depth, $\gamma$, $\lambda$, number of trees (estimators), min child weight, percentage of features, etc with the cross validation applied.
\item Drop a new record all the way down to an external node for all estimators. Calculate the total of the initial prediction and all the corresponding outputs ($w_j$) at external nodes, then set a threshold value to classify the new record. Note that for multi-class classification problem, we'll build $n$ estimators for each class and transform the predictions with softmax function to produce probabilities.
\item  Obtain $$\hat{p}_{new,\,c} = \hat{f}_{xgb,\;c}(X=x_{new}),\;\,\text{for}\;c = 1,\ldots,m$$  and classify $x_{new}$ as $$\hat{y}_{new} = \arg\max_{c=1,\ldots,m}\{\hat{p}_{new,\,c}\}$$

\end{enumerate}
 
\section{Artificial Neural Network (ANN)}

\noindent 
A neural network \cite{Jain} consists of several layers and each layer is made of several units which are known as neurons. Below is the figure of a typical fully connected ANN, which is also named as fully connected feedforward neural network. 
		
		\begin{figure}[htbp]
		\centering
		\includegraphics[scale=0.5]{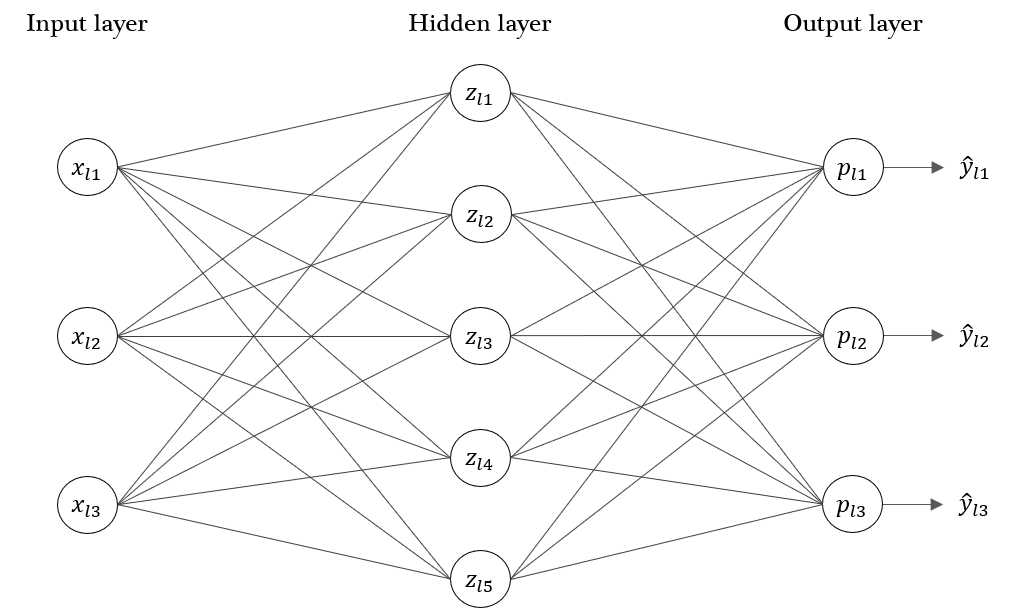}
		\caption{Fully Connected Feedforward Neural Network}
		\label{fig:figure4}
		\end{figure}

\noindent		
The input $\{x_l\}_{l=1,\ldots,n}$ is transmitted from the input layer to the hidden layer. The hidden layer activates the weighted input and output the activated value. The output layer receives the weighted input from hidden layer and activates again to obtain the final $\hat{y}_l$.  Each line in the above figure represents a weight $w_{ij}$ being a multiplier of the corresponding neuron, output layer can be obtained from the following equation (note that Figure \ref{fig:figure4} is drawn under $m=3$).
$$p_{lc}(x_l)  = \sigma_{hidden}(b'_c + \sum^{n_{hidden}}_{j=1}w'_{jc}z_{lj} (\sigma_{input}(b_j + \sum^{n_{input}}_{i=1}w_{ij}x_{li})),\;\text{where}\;c = 1,\ldots,m$$
where $\sigma(\cdot)$ is a non-linear activation function such as the sigmoid function, softmax function, or rectified linear unit and $w, w', b, b'$ are the parameters needed to be estimated. As for how to learn parameters, a loss function has to be specified, we then update the parameters by conducting optimization procedure such as gradient descent to find the optimal parameter combinations.

\medskip\noindent
Learning Steps:
		\begin{enumerate}
		\item Build network structure (determine the number of layers, the number of neurons in each layers, and the type of activation functions for each layers).
		\item Specify a loss function, typically, the cross entropy loss is applied for classification problem. It has the following form.
		$$L = -\sum^m_{c=1}\sum^n_{l=1}y_{lc}\log_2(p_{lc})$$
		\item Update weights $W$ and biases $B$ through gradient descent algorithm with backpropagation applied. Note that a suitable learning rate for training is critical to minimize the loss. 
		\item   Obtain $$\hat{p}_{new,\,c} = \hat{f}_{nn,\;c}(X=x_{new} | W^*, B^*),\;\,\text{for}\;c = 1,\ldots,m$$  and classify $x_{new}$ as $$\hat{y}_{new} = \arg\max_{c=1,\ldots,m}\{\hat{p}_{new,\,c}\}$$
		\end{enumerate}

\chapter{Quadratic Multiform Separation}

\noindent 
In this chapter, we provide a brief description for Quadratic Multiform
Separation, a classification method in machine learning proposed by
Michael Fan et al. in 2019. Please refer to \cite{FanJanuary142021} and \cite{FanApril142021} for further details.

\section{Introduction}

\noindent
Let $\Omega\subset \R^p$ be a collection of observations which is composed of elements of $m$ different memberships. The memberships are labeled as $1,\ldots,m$. A part of observations $\Omega_{\rm tr}\subset\Omega$, typically called training set, and another part of observations $\Omega_{\rm tt}\subset \Omega$, typically called test set, are prepared from $\Omega$. It is assumed that $\Omega_{\tr}$ and $\Omega_{\tt}$  are sufficiently large and these two sets share the same characteristics represented by the whole data space $\Omega$.

Let $\M=\{1,\dots,m\}$ denote the set of possible memberships and $y:\Omega\to\M$ be the membership function that $y(x)$ gives precisely the genuine membership of $x$. The goal of the classification problem is to utilize the training set $\Omega_{\tr}$ to come up with a classifier $\hat{y}(\cdot)$ that serves as a good approximation of $y(\cdot)$. Clearly $y(\cdot)$ and $\hat{y}(\cdot)$ respectively produce a decompositions of $\Omega$ and $\Omega_{\tr}$ as disjoint union of subsets: 
\begin{align}
\Omega &= \bigcup_{j=1}^m \Omega(j) 
= \bigcup_{j=1}^m \hat{\Omega}(j), \\
\Omega_{\tr} &= \bigcup_{j=1}^m \Omega_{\tr}(j) 
= \bigcup_{j=1}^m \hat{\Omega}_{\tr}(j),
\end{align}
where, for all $j=1,\dots,m$,
\begin{align*}
\Omega(j) = \left\{x\in\Omega\,:\,y(x)=j\right\},\;
\Omega_{\tr}(j)=\left\{x\in\Omega_{\tr}:y(x)=j\right\}
\end{align*}
and
\begin{align*}
\hat\Omega(j)  =\left\{x\in\Omega\,:\,\hat{y}(x)=j\right\},\;
\hat\Omega_{\tr}(j)=\left\{x\in\Omega_{\tr}:\hat{y}(x)=j\right\}
\end{align*}
Define $n_{\tr}=|\Omega_{\tr}|$ and $n_{\tr}(j)=|\Omega_{\tr}(j)|$,
where, for a finite set $A$, $|A|$ is the cardinality of the set
$A$, namely, the number of the elements of $A$.
Since the subsets $\Omega_{\tr}(j)$'s are
disjoint and their union is $\Omega_{\tr}$, one has
$n_{\tr}=\sum_{j=1}^m n_{\tr}(j)$.

\section{Multiform Separation}
\subsection{Member Functions}

\noindent 
Let $x=(x_1, \dots, x_p)^T\in \R^p$ and
\begin{equation}
f(x)=\sum_{1\le i\le j \le p} a_{i j}x_i x_j 
+ \sum_{i=1}^p a_i x_i +a_0
\label{f}
\end{equation}
be an arbitrary real quadratic polynomial. The function $f$ is said to be {\it positive} if $f(x)>0$ for all $x\in\R^p$, and {\it nonnegative} if $f(x)\ge 0$ for all $x\in \R^p$. A simple fact is stated below.

\begin{customlemma}{3.1}
Let $f$ be given in (\ref{f}). Then there exists $q\in\N$, constant matrices $A_1, A_2 \in \R^{q\times p}$, and constant vectors $b_1,b_2\in\R^q$ such that any quadratic polynomial $f$ can be expressed as
\begin{equation}
f(x)= \|A_1x-b_1 \|^2-\|A_2x-b_2\|^2
\end{equation}
where $\|\cdot\|$ denotes the Euclidean norm.
Moreover, if $f$ is nonnegative, then
$f(x)= \|Ax-b \|^2$ for some $q,A$, and $b$.
\end{customlemma}

\noindent
{\bf Definition 3.1.} 
A function $f(x):\R^p\to\R$ is said to
be a $q$-{\it dimensional} {\it member function}
if it can be written as
$$f(x) = \|Ax-b\|^2$$
for some integer $q$, $A\in\R^{q\times p}$
and $b\in\R^q$.
Therefore, $f(x)$ is a quadratic function of $x$.
The set of all $q$-dimensional member functions
is denoted by $\Theta(q)$, or simply $\Theta$
when the dimension $q$ is obvious from the context.

Lemma 3.1 implies that any nonnegative quadratic
polynomial in $p$ variables is a $q$-dimensional
member function for some $q\le p+1$.

\subsection{Multiform Separation}

\noindent 
Recall that $\M=\{1,\dots,m\}$ is the set of all possible
memberships, and $\Omega_{\tr}\subset \Omega$ denotes the training set.
Consider $m$ piecewise continuous functions 
$f_j\,:\R^p\,\to\,\R,~j=1,\dots,m$.
Define the 
classifier $\hat{y}\,:\,\Omega\,\to\,\M$ by
\begin{equation}
\hat{y}(x)=j \;\,{\rm if}\;\,
f_j(x) =\min\big\{f_1(x),\dots,f_m(x)\big\}.
\label{MS}
\end{equation}
In other words, an element $x\in \Omega$ is classified by $\hat{y}$ to
have membership $j$ if the evaluation of $f_j$ at $x$ 
is minimal among all $f_k$'s.

\noindent
{\bf Definition 3.2.}
A collection of sets $\{\Lambda_1,\dots,\Lambda_m\}$
is said to be {\it multiform separable}
if there exist piecewise continuous functions $f_1(\cdot),\dots,f_m(\cdot)$
such that for all $i\in\M$ and $x\in\Lambda_i$, 
the condition $f_i(x)\le f_j(x)$ holds for all $j\neq i$.

\subsection{Quadratic Multiform Separation}

\noindent
{\bf Definition 3.3.}
A collection of sets 
$\{\Lambda_1,\dots,\Lambda_m\}$
is said to be {\it quadratic multiform separable}
(or QM-separable) if 
there exist member functions $f_1(\cdot),\dots,f_m(\cdot)$
such that for all $i\in \M$ and $x\in\Lambda_i$, 
the condition $f_i(x)\le f_j(x)$ holds for all $j\neq i$.

\section{Classification Using QMS}

\subsection{Learning Structure}

\noindent 
Given an observation input $x\in\Omega_{\tr}$,
instead of finding a single inferred function $f(x)$ 
which generates a label $i\in\M$ in some optimal sense,
a learning structure which makes use of multiple
inferred functions is given as follows.

\noindent
{\bf Learning Structure.} Given the training set
$\Omega_\tr$ with the corresponding partition
$\Omega_\tr(1),\ldots,$ $\Omega_\tr(m)$, determine whether
the collection of sets $\{\Omega_\tr(1),\dots,\Omega_\tr(m)\}$ is QM-separable.
Alternatively, find member functions
$f_1(\cdot),\dots,f_m(\cdot)$ in such a way that
the collection of sets $\{\Omega_\tr(1),\ldots,\Omega_\tr(m)\}$ is QM-separable
or {\it nearly} QM-separable with respect to
the member functions $f_1(\cdot),\ldots,f_m(\cdot)$.

In the above learning structure,
each set $\Omega_\tr(i)$ is associated with a member function
$f_i(\cdot)$. It then requires that,
given $x\in \Omega_\tr$, the member function
$f_i(x)$ reaches the
lowest value among all member functions
evaluated at the same $x$ if, and only if, 
$x$ belongs to $\Omega_\tr(i)$.
Notice that the associations between the set 
$\Omega_\tr(i)$ and a member
function $f_i$, $i\in\M$, are all correlated.

\subsection{Learning Algorithm}

\noindent 
In this section, the method for constructing
member functions which makes the collection of sets
$$\{\Omega_\tr(1),\dots,\Omega_\tr(m)\}$$
QM-separable is presented. Let $\alpha_{jk}\in [0,1),j=1,\ldots,m,k=1,\ldots,m$ be control hyperparameters of the learning process.
Let $f_1,\dots,f_m$ be $q$-dimensional member functions.
Define the functions 
$\phi_{jk}\,:\,\Omega\,\to\,\R,~j=1,\dots,m,~k=1,\dots,m$, by
\begin{equation}
\phi_{jk}(x)=\max\left\{\alpha_{jk}, 
\frac{f_j(x)}{f_k(x)}\right\}
\label{jk}
\end{equation}

\noindent
It is easy to see that $\phi_{jk}(x) <1 $ if and only if $f_j(x) < f_k(x), j, k\in \M,k\ne j$.
The goal of the learning process is to match
the property ``$x$ has membership $j$" (for some $j\in\M$) 
with the algebraic relations ``$\phi_{jk}(x) <1, k\in\M,k\ne j$".

For $j\in\M$ define $c(j)=c(j; f_1, \dots, f_m)$ by
\begin{equation}
c(j)=c(j; f_1, \dots, f_m)
=\sum_{x\in \Omega_{\tr}(j)} \sum_{k\in\M, k\ne j}
\phi_{jk}(x)  \label{cj}
\end{equation}
$c(j)$ represents the cost contributed by 
$\Omega_{\tr}(j), j\in\M$. 
Denote 
\begin{equation}
\Phi(f_1, \dots, f_m)=\sum_{j=1}^m
c(j; f_1, \dots, f_m)
\label{Phi}
\end{equation}

\noindent
The quantity $\Phi$ provides a performance measure for separating
the sets $\Omega_{\tr}(1), \dots, \Omega_{\tr}(m)$ by the given 
member functions.
With $q$, $\alpha_{jk}$ and $\Omega_{\tr}$ given, 
and $q$ sufficiently large,
the function $\Phi$ defined in (\ref{Phi}) therefore depends only
on $A$'s and $b$'s that define the member functions.

\noindent
{\bf Learning Algorithm.}
Member functions are obtained by minimizing 
$\Phi$ among $\Theta(q)$, 
the set of all $q$-dimensional member functions.
Formally, the task is to solve

\begin{equation}
\min_{f_1, \dots, f_m\in \Theta(q)} \Phi (f_1, \dots, f_m) =
\min_{f_1, \dots, f_m\in \Theta(q)}
~\sum_{j=1}^m \left\{ \sum_{x\in\Omega_{\tr}(j)}
\sum_{k\in\M,k\ne j}
\phi_{jk}(x) 
\right\}
\end{equation}

\noindent
The use of $\alpha_{jk}$ in (\ref{jk}) is to lighten the importance
of $f_j(\cdot)$ at $x$ during the minimization process when
$f_j(x)$ is sufficiently smaller than $f_k(x)$ for all $j\neq k$.

\section{Schematic Diagram}

\noindent 
After previous elaboration of QMS structure, let's have a look on its schematic diagram. This diagram is constructed under the settings of $p = 4, q = 2, m = 3$ meaning that the dataset is $4$-dimensional with $3$ labels, $A_i$ contains $4\times 2$ parameters, and $b_i$ contains $2$ parameters.

	\begin{figure}[htbp]
	\centering
	\includegraphics[scale=0.4]{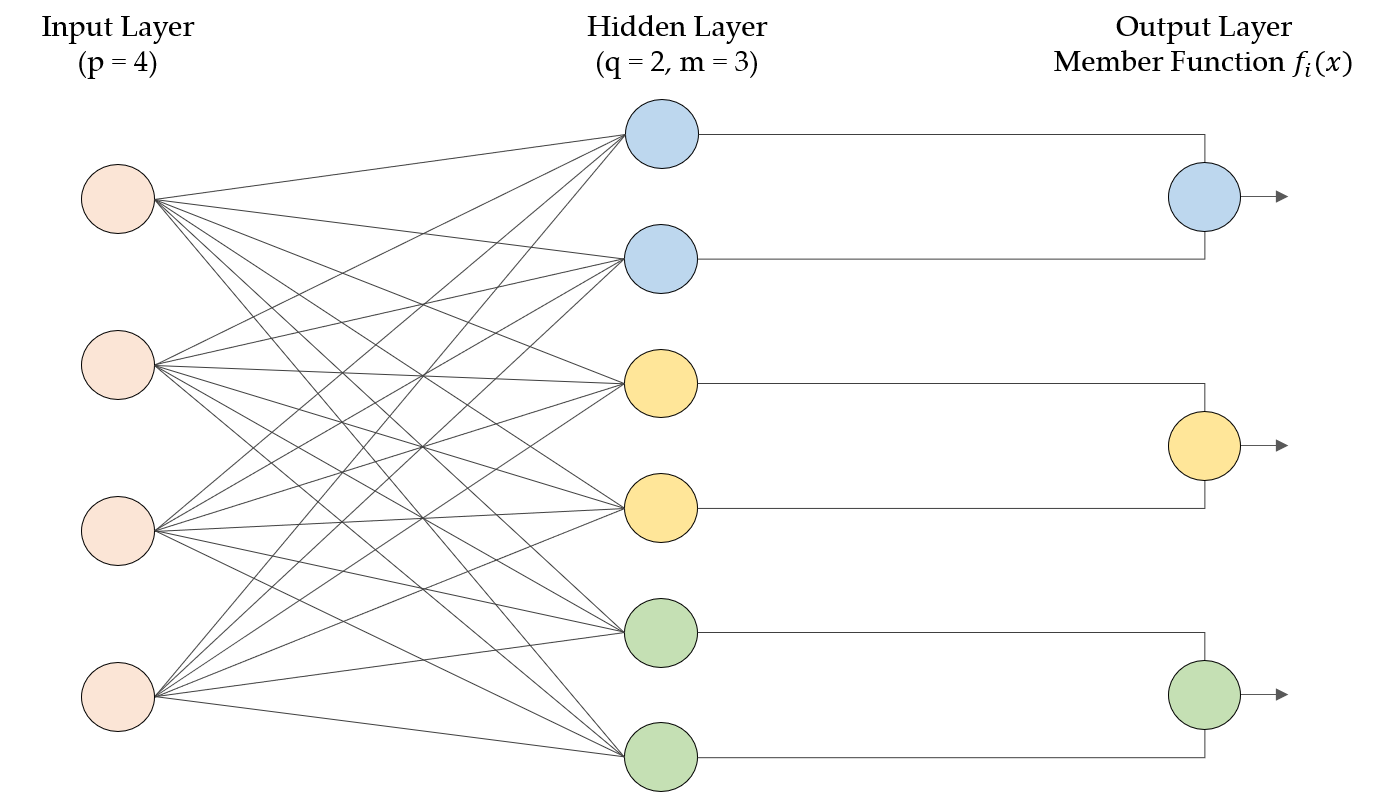}
	\caption{Schematic Diagram of QMS (p = 4, q = 2, m = 3)}
	\label{fig:figure7}
	\end{figure}

\noindent Each line extending from the input layer indicates a weight $a^{(i)}_{jk}$ in $A_i$, where $i\in M, j = 1,\ldots q, k = 1,\ldots,p$, thus, we can see there are $24$ lines equalling to  $pqm = 4\times 2\times 3$. Note that the weights in $b_i$ are not shown in the diagram. Furthermore, the straight lines connect to the output layer indicating a sum of squares operation which allows us to obtain the member function $f_i(x)$ for all $i\in M$. Therefore, it is clear that the number of parameters in QMS is controlled by $p,q,m$, and $q$ is the only hyperparameter that can be specified manually.

\chapter{QMS Optimization and Properties}
\section{Optimization Algorithm}
\noindent 
We propose to train QMS / update the weights of QMS via gradient descent with the technique of Adam\cite{kingma2017adam} applied. Let's recall the form of loss function $\Phi$.
\begin{align*}
\Phi(f_1,\ldots,f_m) 
&= \sum_{j=1}^{m}\left\{\sum_{x\in\,\Omega_{tr}(j)} \sum_{k\in M,\,k\neq j}\max\left\{\alpha_{jk}, \frac{f_j(x)}{f_k(x)}\right\}\right\}\\ 
&= \sum_{j=1}^{m}\left\{\sum_{x\in\,\Omega_{tr}(j)} \sum_{k\in M,\,k\neq j}
\max\left\{\alpha_{jk}, \frac{\twonorm{A_jx}{-}{b_j}^2}{\twonorm{A_kx}{-}{b_k}^2}\right\}\right\}
\end{align*}
where $f_i(x) = \twonorm{A_ix}{-}{b_i}^2$, $A_i\in \R^{q\times p}, b_i\in \R^q, x\in \R^p$ for all $i\in M$, $q$ is a hyperparameter, $p$ denotes the number of attributes, and $m$ denotes the number of classes, where $p$ and $m$ are predetermined by data. Our goal is to find a classifier with the best combinations of $\{A_i, b_i\}_{i\in M}$ by minimizing $\Phi(f_1,\ldots,f_m)$. Below we demonstrate a simple case for above equation under $m=2$ to get insights of it.
\begin{align*}
\Phi(f_1, f_2) & = \sum_{x\in\Omega_{tr}(1)}\max\left\{\alpha_{12},\frac{f_1(x)}{f_2(x)}\right\} + \sum_{x\in\Omega_{tr}(2)}\max\left\{\alpha_{21},\frac{f_2(x)}{f_1(x)}\right\} 
\\& = \sum_{x\in\Omega_{tr}(1)}\max\left\{\alpha_{12},\frac{\twonorm{A_1x}{-}{b_1}^2}{\twonorm{A_2x}{-}{b_2}^2}\right\} + \sum_{x\in\Omega_{tr}(2)}\max\left\{\alpha_{21}, \frac{\twonorm{A_2x}{-}{b_2}^2}{\twonorm{A_1x}{-}{b_1}^2}\right\}
\end{align*}
As we can find from above example, $A_1, A_2, b_1$ and $b_2$ are the parameter matrices required to be updated. Thus, it is clear that the total number of parameter matrices is $m\times 2$ with the actual number of parameters $qm(p + 1)$ which can be derived from the following formula.
$$
\text{number of parameters} = qpm + qm = qm(p + 1)
$$
Therefore, under the case of $m=2$, there are $4$ parameter matrices with $2q(p+1)$ total parameters. After revealing the unknown parameters in QMS-specific loss function, it's time to deep dive into the mathematical derivation and the training algorithm of QMS.
\subsection{Mathematical Derivation of Gradients}

\noindent 
Before entering the gradient calculation, let's present several lemmas regarding matrix calculus that will be used subsequently.
\begin{customlemma}{4.1}
Given $\boldsymbol{a} \in \R^q, \boldsymbol{b} \in \R^p$ and $X\in \R^{q\times p}$, where $\boldsymbol{a}, \boldsymbol{b}$ are not functions of $X$, we have the following equalities.
\begin{align}
\mathrlap{\pdv{\boldsymbol{a}^TX\boldsymbol{b}}{X}}\phantom{xxxxxxx} = \mathrlap{\pdv{\boldsymbol{b}^TX^T\boldsymbol{a}}{X}}\phantom{xxxxxxx} = \boldsymbol{ab}^T
\end{align}
\label{l1}
\end{customlemma}
\begin{proof}
It is straightforward to show that the first equality in (4.1) holds since $(\boldsymbol{a}^TX\boldsymbol{b})^T = \boldsymbol{b}^TX^T\boldsymbol{a}$. For any given $i \in \{1,2,\ldots,q\}, j\in \{1,2,\ldots,p\}$, we can derive
\begin{align*}
\left(\pdv{\boldsymbol{a}^TX\boldsymbol{b}}{X}\right)_{ij} & = \pdv{\sum\limits^q_{k=1}\sum\limits^p_{l=1} a_k x_{kl}b_l}{x_{ij}}\\
& = \sum_{k=1}^q\sum_{l=1}^p a_kb_l\pdv{x_{kl}}{x_{ij}} \\
& = \sum_{k=1}^q\sum_{l=1}^p a_kb_l\delta_{ik}\delta_{jl} \\
& = a_ib_j  \\
& = (\boldsymbol{ab}^T)_{ij}
\end{align*}
where $\delta_{ij} = 1$ if $i=j$ and $0$ otherwise, and thus (4.1) holds.
\end{proof}
\clearpage
\begin{customlemma}{4.2}
Given $\boldsymbol{a}, \boldsymbol{b} \in \R^p$ and $X\in \R^{q\times p}$ where $\boldsymbol{a}, \boldsymbol{b}$ are not functions of $X$, the following equation holds.
\begin{align}
\pdv{\boldsymbol{a}^TX^TX\boldsymbol{b}}{X} = X(\boldsymbol{ab}^T+\boldsymbol{ba}^T)
\end{align} 
\begin{proof}
{\savebox\strutbox{$\vphantom{\dfrac11}$}
\begin{align*}
\pdv{\boldsymbol{a}^TX^TX\boldsymbol{b}}{X}
& = \left.\left(\pdv{\boldsymbol{a}^TY^TX\boldsymbol{b}}{X} + \pdv{\boldsymbol{a}^TX^TY\boldsymbol{b}}{X}\right)\right|_{Y = X} \\
& =  \left.\left(\pdv{(Y\boldsymbol{a})^TX\boldsymbol{b}}{X} + \pdv{\boldsymbol{a}^TX^T(Y\boldsymbol{b})}{X}\right)\right|_{Y = X}\\
& = X\boldsymbol{a}\boldsymbol{b}^T + X\boldsymbol{b}\boldsymbol{a}^T \\
& = X(\boldsymbol{a}\boldsymbol{b}^T + \boldsymbol{b}\boldsymbol{a}^T)
\end{align*}}
\end{proof}
\label{l2}
\end{customlemma}
\noindent Next, solving the gradients $\Phi(f_1,\ldots,f_m)$ w.r.t. $A_i$ and $b_i$ requires the calculations of partial $f_i(x)$ over partial $A_i$ and $b_i$ for all $i \in M$; therefore, let's structure a lemma for them first based on Lemmas \ref{l1} and \ref{l2}.
\begin{customlemma}{4.3}
Given $x \in \R^p$, $A_i\in \R^{q\times p}$, $b_i \in \R^{q}$, and $f_i(x) = \twonorm{A_ix}{-}{b_i}^2$, where  $x$ is not a function of $A_i$ or $b_i$ for all $i\in M$. The following equation holds.
\begin{align}
\pdv{f_i(x)}{A_i} &= 2(A_ix - b_i)x^T \\[8pt]
\pdv{f_i(x)}{b_i} &= -2(A_ix - b_i)
\end{align}
\label{l3}
\end{customlemma}
\begin{proof}
We'll first prove (4.3), then (4.4).
{\savebox\strutbox{$\vphantom{\dfrac11}$}
\begin{align*}
\pdv{f_i(x)}{A_i}&= \pdv{A_i}\twonorm{A_ix}{-}{b_i}^2 \\
&= \pdv{A_i}(A_ix-b_i)^T(A_ix-b_i)\\
&= \pdv{A_i}(x^TA_i^TA_ix -x^TA^T_ib_i-b_i^TA_ix+b_i^Tb_i)\\
&= A_i(xx^T + xx^T) -2b_ix^T \\
&= 2(A_ixx^T - b_ix^T)\\
& = 2(A_ix - b_i)x^T \\[8pt]
\pdv{f_i(x)}{b_i} &= \pdv{b_i}\twonorm{A_ix}{-}{b_i}^2 \\
&= \pdv{b_i}(A_ix-b_i)^T(A_ix-b_i) \\
&= \pdv{b_i}(x^TA_i^TA_ix -x^TA^T_ib_i-b_i^TA_ix+b_i^Tb_i) \\
&= -2(A_ix - b_i)
\end{align*}}
\end{proof}
\noindent After achieving partial derivatives of $f_i(x)$ with respect to $A_i$ and $b_i$ for all $i \in M$, we are one step away from finalizing the gradient of the loss function $\Phi(f_1(x),\ldots,f_m(x))$. Since there exists $\max$ functions in $\Phi(f_1(x),\ldots,f_m(x))$ being not differentiable everywhere, we consider to include the singularity (turning point) as a part of horizontal line in $\max$ function, which is the convention borrowed from using gradient descent to train neural network with ReLU (Rectified Linear Unit) activation function applied. Note that the ReLU activation function has the following form.
$$
g(z) =\max\{0,z\},\;\text{where}\;\,z\in \R
$$
The derivative of $g(z)$ w.r.t. $z$ according to the above-mentioned convention is
\begin{equation*}
\dv{g(z)}{z} = 
\begin{cases}
1, & \text{if}\ z > 0 \\
0, & \text{otherwise}
\end{cases}
\end{equation*}

\begin{customthm}{4.1} The gradients of the QMS's loss function w.r.t. $A_i$ and $b_i$ are given by 
{\savebox\strutbox{$\vphantom{\dfrac11}$}
\begin{align}
\mathrlap{\nabla \Phi(A_i)}\phantom{xxxxxX} & = 2\sum_{j\in M,\,j\neq i}\left(\sum_{x\in \Omega_{tr}(i)} I_{ij}(x)\frac{(A_ix-b_i)x^T}{f_j(x)} - \sum_{x\in \Omega_{tr}(j)}I_{ji}(x)\frac{f_j(x)(A_ix-b_i)x^T}{f_i(x)^2}\right)\\[8pt]
\mathrlap{\nabla \Phi(b_i)}\phantom{xxxxxX} & = -2\sum_{j\in M,\,j\neq i}\left(\sum_{x\in \Omega_{tr}(i)} I_{ij}(x) \frac{A_ix-b_i}{f_j(x)} - \sum_{x\in \Omega_{tr}(j)}I_{ji}(x)\frac{f_j(x)(A_ix-b_i)}{f_i(x)^2}\right)   
\end{align}}
where 
\begin{equation*}
I_{ij}(x) = I\left(\frac{f_i(x)}{f_j(x)} > \alpha_{ij}\right) = 
\begin{cases}
1, & \text{if}\ \frac{f_i(x)}{f_j(x)} >  \alpha_{ij}\\
0, & \text{otherwise}
\end{cases}
\end{equation*}
\label{l4}
\end{customthm}
\begin{proof}
Let $Z_{ij}$ denote $\dfrac{f_i{(x)}}{f_j{(x)}}$, where $i$ is the label in focus, $i, j \in M$ and $i\neq j$, we have
\begin{align*}
\mathrlap{\nabla \Phi(A_i)}\phantom{xxxxxX}  &= \pdv{\Phi(f_1,\ldots,f_m)}{A_i}\\
& = \sum_{j\in M,\,j\neq i} \sum_{x\in \Omega_{tr}(i)}\pdv{\max\{\alpha_{ij}, Z_{ij}\}}{Z_{ij}}\pdv{Z_{ij}}{A_i} +  \sum_{j\in M,\,j\neq i}\sum_{x\in \Omega_{tr}(j)}\pdv{\max\{\alpha_{ji}, Z_{ji}\}}{Z_{ji}}\pdv{Z_{ji}}{A_i}\\
& = \sum_{j\in M,\,j\neq i}\left(\sum_{x\in \Omega_{tr}(i)}I_{ij}(x)\pdv{A_i}\frac{\twonorm{A_ix}{-}{b_i}^2}{f_j(x)} +  \sum_{x\in \Omega_{tr}(j)}I_{ji}(x)\pdv{A_i} \frac{f_j(x)}{\twonorm{A_ix}{-}{b_i}^2}\right)\\
& =  2\sum_{j\in M,\,j\neq i}\left(\sum_{x\in \Omega_{tr}(i)} I_{ij}(x)\frac{(A_ix-b_i)x^T}{f_j(x)} - \sum_{x\in \Omega_{tr}(j)}I_{ji}(x)\frac{f_j(x)(A_ix-b_i)x^T}{f_i(x)^2}\right)\\[8pt]
\mathrlap{\nabla \Phi(b_i)}\phantom{xxxxxX}  &= \pdv{\Phi(f_1,\ldots,f_m)}{b_i}\\
& = \sum_{j\in M,\,j\neq i} \sum_{x\in \Omega_{tr}(i)}\pdv{\max\{\alpha_{ij}, Z_{ij}\}}{Z_{ij}}\pdv{Z_{ij}}{b_i} +  \sum_{j\in M,\,j\neq i}\sum_{x\in \Omega_{tr}(j)}\pdv{\max\{\alpha_{ji}, Z_{ji}\}}{Z_{ji}}\pdv{Z_{ji}}{b_i}\\
& = \sum_{j\in M,\,j\neq i} \left(\sum_{x\in \Omega_{tr}(i)}I_{ij}(x)\pdv{b_i}\frac{\twonorm{A_ix}{-}{b_i}^2}{f_j(x)} +  \sum_{x\in \Omega_{tr}(j)}I_{ji}(x)\pdv{b_i} \frac{f_j(x)}{\twonorm{A_ix}{-}{b_i}^2}\right)\\
& = -2\sum_{j\in M,\,j\neq i}\left(\sum_{x\in \Omega_{tr}(i)} I_{ij}(x) \frac{A_ix-b_i}{f_j(x)} - \sum_{x\in \Omega_{tr}(j)}I_{ji}(x)\frac{f_j(x)(A_ix-b_i)}{f_i(x)^2}\right)   
\end{align*}
\end{proof}

\noindent After above mathematical derivations, we then obtain the gradient of $\Phi(f_1,\ldots,f_m)$ with respect to $A_i$ and $b_i$. In the sequel, Theorem \ref{l4} will serve as descent directions for updating the parameters in matrices $A_i, b_i$ for all $i\in M$.

\clearpage
\subsection{Training Algorithm and Results}

\noindent 
The section is divided into 2 parts, which are the procedure of Adam and the computations of Member Function and Loss Function. Adam (Adaptive Moment Estimation) is a gradient descent method introduced by Diederik P. Kingma and Jimmy Lei Ba in 2015\cite{kingma2017adam}. The following paragraph is an excerpt from their paper $-$ Adam: A Method for Stochastic Optimization.  
\begin{displayquote}
Adam is an algorithm for first-order gradient-based optimization of stochastic objective functions, based on adaptive estimates of lower-order moments. The method is straightforward to implement, is computationally efficient, has little memory requirements, is invariant to diagonal rescaling of the gradients, and is well suited for problems that are large in terms of data and/or parameters. The method is also appropriate for non-stationary objectives and problems with very noisy and/or sparse gradients. The hyper-parameters have intuitive interpretations and typically require little tuning.
\end{displayquote}

Moreover, Adam is known as a combination of Momentum and RMSprop. Momentum helps the update not only to move on when it meets the saddle point, but also to render it escape from the local minimum by pushing it across the hill of the loss function. RMSprop serves as a brake or an accelerator when the gradients are too large or small. Momentum and RMSprop have the following forms,
\begin{align*}
w^{t+1}_i &\gets w^t_i+v_i^t,\;\text{where}\;v_i^0 = 0,\;\text{and}\;v_i^t = \lambda v_i^{t-1}-\eta g_i^t &\text{(Momentum)}\\
w^{t+1}_i &\gets w^t_i - \frac{\eta}{\sigma_i^t}g_i^t,\;\text{where}\;\sigma_i^0=\sqrt{(g_i^0)^2},\; \text{and}\;\sigma_i^t=\sqrt{\alpha(\sigma_i^{t-1})^2 + (1-\alpha)(g_i^t)^2} &\text{(RMSprop)}
\end{align*}
where $w$ is a weight vector, $g_i^t = \left.\pdv{L}{w_i}\right|_{w=w^t}$, and $\lambda$ in Momentum and $\alpha$ in RMSprop are hyperparameters to control the extent of current gradient's contribution for updating $w$. 

Now, let's incorporate Adam into QMS. We denote $\boldsymbol{A}$ and $\boldsymbol{b}$ as sets of $A_i$ and $b_i$ for all $i\in M$, respectively. Since we have already obtained the gradients of $\Phi(f_1,\ldots,f_m)$ w.r.t. $\boldsymbol{A}$ and $\boldsymbol{b}$ in the previous section. The pseudo code of QMS + Adam training algorithm is presented in the following page.
\begin{algorithm}[H]
\caption{QMS + Adam Training Procedure\\
A good default setting for learning rate is $lr_A = lr_b = 1$. As For Adam-related hyperparameters, the default settings follow original paper which are  $\beta_{A1} = \beta_{b1} = 0.9$, $\beta_{A2} = \beta_{b2} = 0.999$, and $\epsilon = 10^{-8}$. Note that $\beta^t_{A1},\beta^t_{b1},\beta^t_{A2},\beta^t_{b2}$ are the corresponding $\beta$ to the power of $t$.}
\begin{spacing}{1.15}
\begin{algorithmic}[1]
\Require Learning rates $lr_A, lr_b$
\Require Exponential decay rates $\beta_{A1}, \beta_{A2}, \beta_{b1}, \beta_{b2} \in [0,1)$
\Require Loss Function $\Phi(f, \alpha)$
\Ensure $A_i \in \R^{q\times p},\; b_i \in \R^{q\times 1}$
\State \textbf{Initialize:}
\State \hspace*{\algorithmicindent}\parbox[t]{.8\linewidth}{$A_i \gets N(0,1),\;b_i \gets N(0,1)$ {\footnotesize (Initialize parameters, element-wise)}}
\State \hspace*{\algorithmicindent}\parbox[t]{.8\linewidth}{$\mathrlap{m_{A_i,\,0}}\phantom{xxxx} \gets 0,\;\mathrlap{m_{b_i,\,0}}\phantom{xxxx} \gets 0$ {\footnotesize  (Initialize $1^{st}$ moment matrix)}}
\State \hspace*{\algorithmicindent}\parbox[t]{.8\linewidth}{$\mathrlap{v_{A_i,\,0}}\phantom{xxxx} \gets 0,\;\mathrlap{v_{b_i,\,0}}\phantom{xxxx} \gets 0$ {\footnotesize  (Initialize $2^{nd}$ moment matrix)}}
\State \hspace*{\algorithmicindent}\parbox[t]{.8\linewidth}{$t\gets 0$ {\footnotesize (Initialize time step)}}
\While{$\boldsymbol{A_t}, \boldsymbol{b_t}$ not converged}
	\State $t \gets t+1$
	\For{$i \in \{1,\ldots,m\}$}	
		\State $\mathrlap{g_{A_i,\,t}}\phantom{xxxx} \gets \nabla_{A_i}\Phi_t(A_{i,\,t-1})$  {\footnotesize (Obtain $A_i$'s gradient)}
		\State $\mathrlap{g_{b_i,\,t}}\phantom{xxxx} \gets \nabla_{b_i}\Phi_t(b_{i,\,t-1})$  {\footnotesize (Obtain $b_i$'s gradient)}
		\State $\mathrlap{m_{A_i,\,t}}\phantom{xxxx} \gets \beta_{A1}\cdot m_{A_i,\,t-1} + (1-\beta_{A1})\cdot g_{A_i,\,t}$  {\footnotesize (Update $A_i$'s biased $1^{st}$ moment estimate)}
		\State $\mathrlap{m_{b_i,\,t}}\phantom{xxxx} \gets \beta_{b1}\cdot m_{b_i,\,t-1} + (1-\beta_{b1})\cdot g_{b_i,\,t}$   {\footnotesize (Update $b_i$'s  biased $1^{st}$ moment estimate)}
		\State $\mathrlap{v_{A_i,\,t}}\phantom{xxxx} \gets \beta_{A2}\cdot v_{A_i,\,t-1} + (1-\beta_{A2})\cdot g^2_{A_i,\,t}$   {\footnotesize (Update $A_i$'s biased $2^{nd}$ raw moment estimate)}
		\State $\mathrlap{v_{b_i,\,t}}\phantom{xxxx} \gets \beta_{b2}\cdot v_{b_i,\,t-1} + (1-\beta_{b2})\cdot g^2_{b_i,\,t}$   {\footnotesize (Update $b_i$'s biased $2^{nd}$ raw moment estimate)}
		\State $\mathrlap{\widehat{m}_{A,\,t}}\phantom{xxxx} \gets m_{A_i,\,t} / (1-\beta^t_{A1})$ {\footnotesize (Compute $A_i$'s bias-corrected $1^{st}$ moment estimate)}
		\State $\mathrlap{\widehat{m}_{b,\,t}}\phantom{xxxx} \gets m_{b_i,\,t} / (1-\beta^t_{b1})$ {\footnotesize (Compute $b_i$'s bias-corrected $1^{st}$ moment estimate)}
		\State $\mathrlap{\widehat{v}_{A,\,t}}\phantom{xxxx} \gets v_{A_i,\,t} / (1-\beta^t_{A2})$  {\footnotesize (Compute  $A_i$'s bias-corrected $2^{nd}$ raw moment estimate)}
		\State $\mathrlap{\widehat{v}_{b,\,t}}\phantom{xxxx} \gets v_{b_i,\,t} / (1-\beta^t_{b2})$ {\footnotesize (Compute  $b_i$'s bias-corrected $2^{nd}$ raw moment estimate)}
		\State $\mathrlap{A_{i,\,t}}\phantom{xxxx} \gets A_{i,\,t-1} - lr_A\cdot \widehat{m}_{A,\,t} / (\sqrt{\widehat{v}_{A,\,t}}+\epsilon)$  {\footnotesize (Update $A_i$)}
		\State $\mathrlap{b_{i,\,t}}\phantom{xxxx} \gets b_{i,\,t-1} - lr_b\cdot \widehat{m}_{b,\,t} / (\sqrt{\widehat{v}_{b,\,t}}+\epsilon)$ {\footnotesize (Update $b_i$)}
	\EndFor
\EndWhile\\
\Return $\boldsymbol{A_t},\;\boldsymbol{b_t}$ {\footnotesize (Resulting parameters)}
\end{algorithmic}
\end{spacing}
\end{algorithm}
\noindent $\Phi(f_1,\ldots,f_m)$ can be a noisy function which the noise derived from the evaluation on mini batches of training data points. In this case, the minimization of the $\Phi$'s expected value $E[\Phi(\boldsymbol{A}, \boldsymbol{b})]$ w.r.t. $\boldsymbol{A}$ and $\boldsymbol{b}$ is of interest. The reason of computing bias-corrected moment estimates is because that the (element-wise) initial value of them are all set as zeros, resulting in the moment estimates are biased towards zeros, especially during the initial time steps, and especially when the decay rates are small.

In the second part, we provide the pseudo codes of member function and loss function. Note that $\Omega_{tr}$  and $\Omega_{tr}(i)$ are denoted as $X$ and $X^{(i)}$, respectively.
\begin{algorithm}[H]
\caption{Member Function}
\begin{spacing}{1.15}
\begin{algorithmic}[1]
\Require $A_i \in \R^{q\times p},\;b_i \in \R^{q\times 1},\; X^{(i)} \in \R^{p\times n_i}$
\Ensure $f(X;\boldsymbol{A},\boldsymbol{b})\in \R^{m\times m\times n_i}$
\Function{Member}{$X$}
\For{$i \in \{1,\ldots,m\}$}
	\For{$j \in \{1,\ldots,m\}$}
		\State $f[i][j] \gets \twonorm{A_j X^{(i)}}{-}{b_j}^2$ {\footnotesize(column-wise euclidean norm)}
	 \EndFor
\EndFor
\State \Return $f$
\EndFunction
\end{algorithmic}
\end{spacing}
\end{algorithm}
\begin{algorithm}[H]
\caption{Loss Function}
\begin{spacing}{1.15}
\begin{algorithmic}[1]
\Require \Call{Member}{$X$}
\Ensure $\Phi(f, \alpha) \in \N$
\Function{Loss}{$f, \alpha$}
\State $\Phi \gets 0$
\For{$i \in \{1,\ldots,m\}$}
	\For{$j \in \{1,\ldots,m\}$}
		\If{$i \neq j$}
			\State $\Phi \gets \Phi + Sum(Max(f[i][i] / f[i][j], \alpha))$
		\EndIf
	 \EndFor
\EndFor
\State \Return $\Phi$ 
\EndFunction
\end{algorithmic}
\end{spacing}
\end{algorithm}
\noindent Once the converged $A_i$ and $b_i$ for all $i\in M$ are obtained, an observation $x \in \Omega$ is classified by $\hat{y}$ to have membership $j$ if the evaluation of $f_j$ at $x$ is minimal among all $f_k$'s.
$$
\hat{y}(x) = j\;\,\text{if}\;\,f_j(x) = \min\{f_1(x),\ldots,f_m(x)\}
$$

\section{Properties of Hyperparameters}

\noindent 
In this section, the exploration between hyperparameters and accuracy is conducted with 10-fold cross validation and evaluated on four datasets MNIST, Fashion MNIST, Dry Bean and Census Income. The introduction of these datasets will be provided in section 5.1.

\subsection{Weight Hyperparameter $q$}

\noindent 
In each member function $f_i(x) = \twonorm{A_ix}{-}{b_i}^2$, where $x \in \R^p, A_i \in \R^{p\times q}, b_i\in \R^{q},\text{and}\; i \in M$. Our goal is to find the optimal $A_i$ and $b_i$ to construct a classifier by minimizing the QMS-specific loss function. As we can see from Section 4.1 that the total number of parameters considered in QMS is $qm(p+1)$ being only depending on $p,q$ and $m$. Since $p$ and $m$ are predetermined by the data, $q$ is the only hyperparameter that is of interest to explore. Note that $p$ denotes the number of attributes, $q$ is the hyperparameter that can be set manually, and $m$ denotes the number of classes.

\begin{figure}[htbp]
\centering
\begin{subfigure}{.4\textwidth}
  \centering
  \includegraphics[width=1\linewidth]{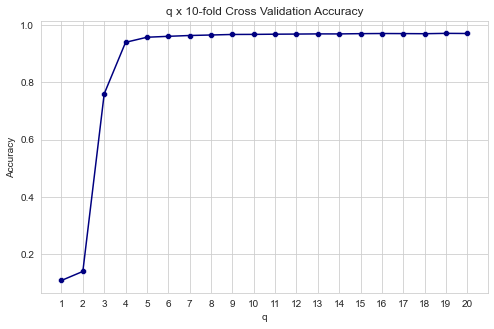}
  \caption{\tiny MNIST (epoch = 10, batch = 200, $\alpha$ = 0.30)}
\end{subfigure}%
\begin{subfigure}{.4\textwidth}
  \centering
  \includegraphics[width=1\linewidth]{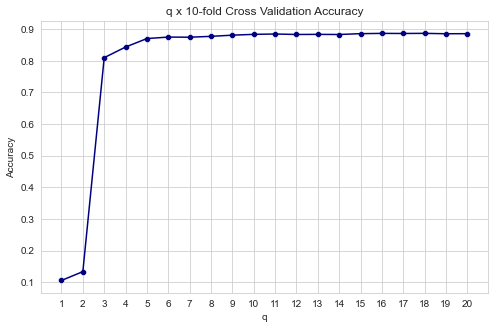}
  \caption{\tiny Fashion MNIST (epoch = 10, batch = 200, $\alpha$ = 0.40)}
\end{subfigure}
\begin{subfigure}{.4\textwidth}
  \centering
  \includegraphics[width=1\linewidth]{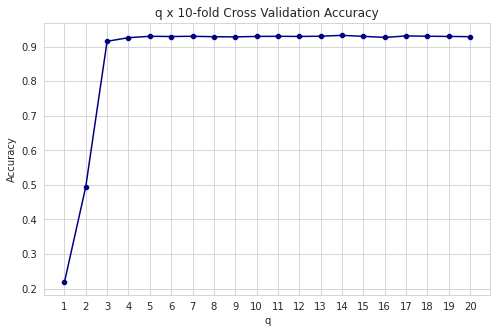}
  \caption{\tiny Dry Bean (epoch = 15, batch = 200, $\alpha$ = 0.40)}
\end{subfigure}%
\centering
\begin{subfigure}{.4\textwidth}
  \centering
  \includegraphics[width=1\linewidth]{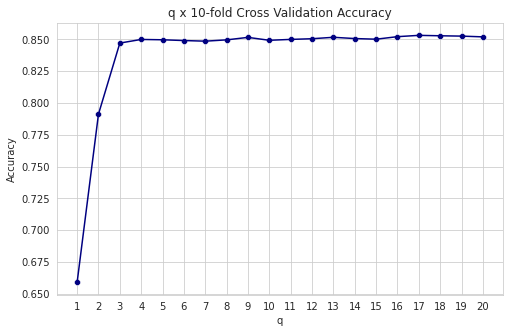}
  \caption{\tiny Census Income (epoch = 6, batch = 1000, $\alpha$ = 0.50)}
\end{subfigure}
\caption{Evaluate $q$ on Different Datasets}
\label{fig:}
\end{figure}

\clearpage
From above figures, we unveil that as $q$ goes up, the accuracy goes higher. This is obvious since there are more parameters utilized in the QMS algorithm. However, the accuracy remains nearly unchanged after the total parameters utilized reach a certain amounts of number. We usually set $q$ as $15$ since in most of the cases, it is sufficiently large to produce a good result.

\subsection{Control Hyperparameter $\alpha$}

\noindent 
Recall that the loss function $\Phi(f_1,\ldots,f_m)$ has the following form.
\begin{equation}
\Phi(f_1,\ldots,f_m) = \sum_{j=1}^{m}\left\{\sum_{x\in\,\Omega_{tr}(j)} \sum_{k\in M,\,k\neq j}\max\left\{\alpha_{jk}, \frac{f_j(x)}{f_k(x)}\right\}\right\}
\label{4.7}
\end{equation}
As we know that the use of $\alpha_{jk}$ in (\ref{4.7}) is to lighten the importance of $f_j(\cdot)$ at $x$ during the minimization process when $f_j(x)$ is sufficiently smaller than $f_k(x)$ for all $j\neq k$. For providing a fundamental understanding, we let $\alpha_{jk} \equiv \alpha,\;\forall j,\,k$. Thus, in the following, we are going to explore how critical the $\alpha$ is.

\begin{figure}[htbp]
\centering
\begin{subfigure}{.4\textwidth}
  \centering
  \includegraphics[width=1\linewidth]{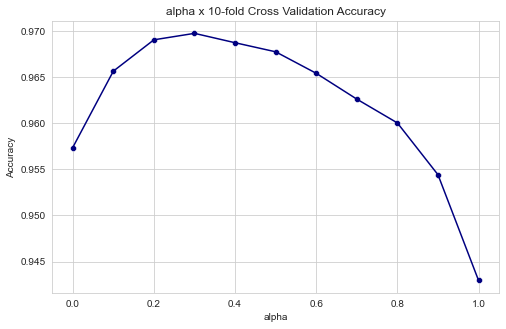}
  \caption{\tiny MNIST ($q = 15$, best $\alpha = 0.30$)}
\end{subfigure}%
\begin{subfigure}{.4\textwidth}
  \centering
  \includegraphics[width=1\linewidth]{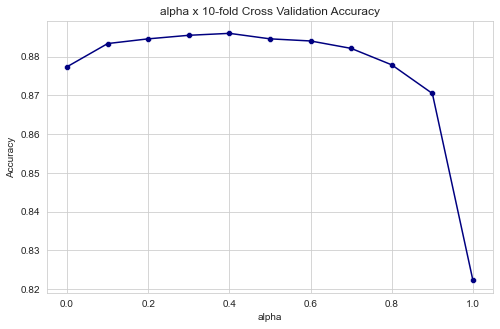}
  \caption{\tiny Fashion MNIST ($q=15$, best $\alpha =  0.40$)}
\end{subfigure}
\begin{subfigure}{.4\textwidth}
  \centering
  \includegraphics[width=1\linewidth]{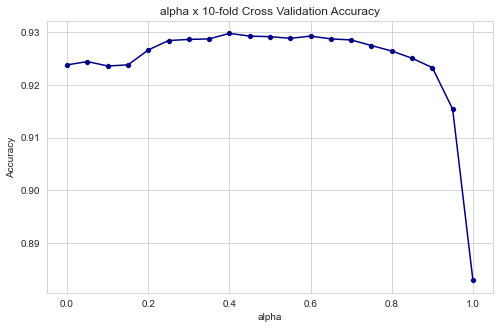}
  \caption{\tiny Dry Bean ($q = 15$, best $\alpha = 0.40$)}
\end{subfigure}%
\begin{subfigure}{.4\textwidth}
  \centering
  \includegraphics[width=1\linewidth]{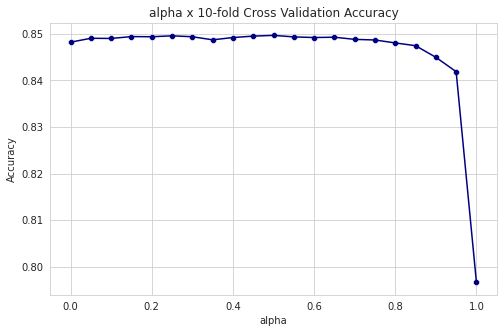}
  \caption{\tiny Census Income ($q = 5$, best $\alpha = 0.50$)}
\end{subfigure}
\caption{Evaluate $\alpha$ on Different Datasets}
\label{fig:}
\end{figure}

As we can see from above figures that $\alpha$ is a tunable hyperparameter with a single peak. The further the alpha is away from the peak, the worse the corresponding accuracy is. We also unveil that in most of the cases, $\alpha$ is insensitive to accuracy when it is set small comparing to the opposite. Furthermore, the best $\alpha$ usually occurs  when $\alpha \leq 0.5$, therefore, it is recommended to not set a large $\alpha$ during model tuning. 

\chapter{Empirical Model Performance}

\noindent 
In this section, QMS is going to be compared with six popular classification methods including k-Nearest Neighbors, Logistic Regression, Support Vector Machine, Random Forest, eXtreme Gradint Boosting, and Artificial Neural Network on five different datasets. These include MNIST, Fashion MNIST, Dogs vs. Cats, Census Income, and Dry Bean. 

\section{Description of Datasets}

\noindent 
Below is the list of datasets considered and the corresponding descriptions.

\begin{table}[htbp]
\caption{Dataset Summary}
\renewcommand{\arraystretch}{1.4} 
\centering
\begin{tabular}{ccccc}
\toprule
Name & Data Type & \# Instances & \# Attributes & Classification Task \\ \midrule
MNIST & Image & 70,000& $28 \times 28$  & Multi-class (10)\\
Fashion MNIST & Image& 70,000 & $28 \times 28$ & Multi-class (10)\\ 
Dogs vs. Cats & Image &  25,000 &  $224 \times 224$ & Binary \\
Census Income & Multivariate  & 45,222  & 14  & Binary\\ 
Dry Bean & Multivariate  & 13,611   & 16 & Multi-class (7)\\ 
\bottomrule
\end{tabular}
	\label{Tab:}
\end{table}

\subsection{MNIST}
\noindent 
The MNIST database\cite{lecun-mnisthandwrittendigit-2010} (Modified National Institute of Standards and Technology database) released in 1999 is a large database of handwritten digits. The task of this dataset is to classify image into correct number. It consists of a training set of 60,000 examples and a test set of 10,000 examples.  Each example is a 28$\times$28 grayscale image, associated with a label from 10 classes.

\begin{figure}[htbp]
\centering
\includegraphics[scale=0.35]{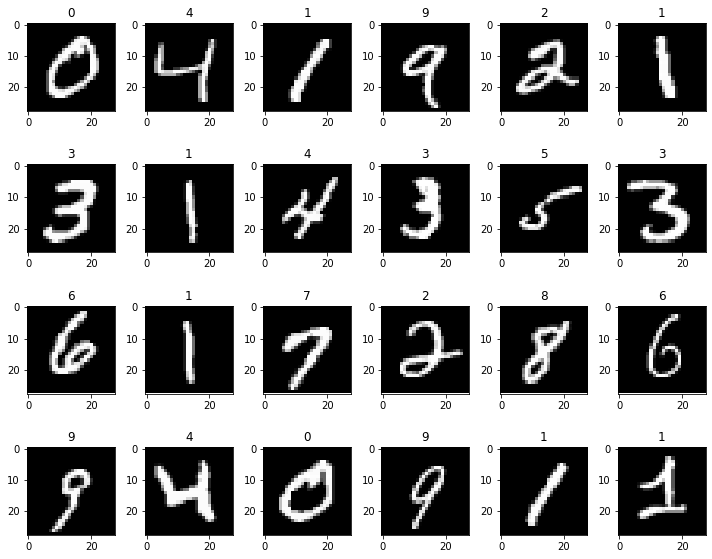}
\caption{MNIST Sample Data}
\label{Fig:}
\end{figure}

\begin{itemize}
\item Labels: \textbf{10 labels}
\begin{table}[htbp]
\caption{MNIST Label Summary}
\renewcommand{\arraystretch}{1.3} 
\centering
\begin{tabular}{|ccc|ccc|}
\hline
Label & Name & \# Instances & Label & Name & \# Instances\\ 
\hhline{|===|===|}
0 & 0 & 6,903 & 5 & 5 & 6,313\\
1 & 1 & 7,877 & 6 & 6 & 6,876\\
2 & 2 & 6,990 & 7 & 7 & 7,293\\
3 & 3 & 7,141 & 8 & 8 & 6,825\\
4 & 4 & 6,824 & 9 & 9 & 6,958\\  
\hline
\end{tabular}
	\label{Tab:}
\end{table}
\item Attributes: \textbf{784 attributes (pixels)}
\end{itemize}

\subsection{Fashion MNIST}
\noindent 
Fashion-MNIST dataset\cite{xiao2017fashionmnist} released in 2017 is a dataset from Zalando's article.  The task of this dataset is to classify image into correct category. It includes training set of 60,000 examples and a test set of 10,000 examples. Each example is a 28$\times$28 grayscale image, associated with a label from 10 classes.

\begin{figure}[htbp]
\renewcommand{\arraystretch}{1.3} 
\centering
\includegraphics[scale=0.35]{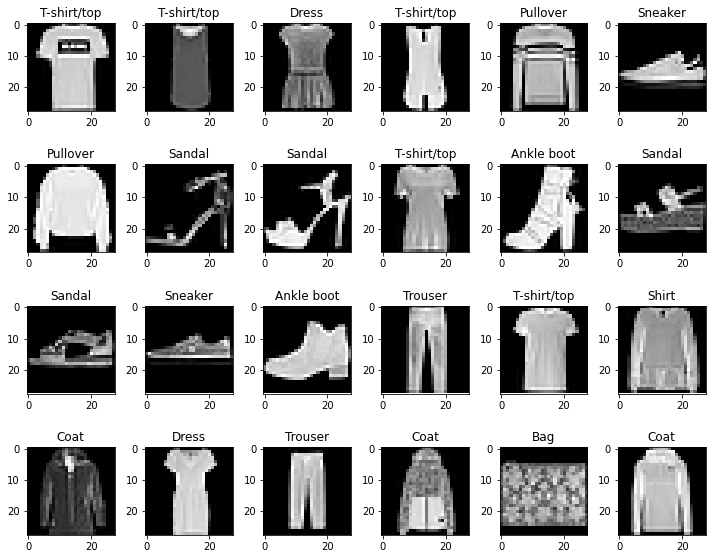}
\caption{Fashion MNIST Sample Data}
\label{Fig:}
\end{figure}

\begin{itemize}
\item Labels: \textbf{10 labels}
\begin{table}[htbp]
\caption{Fashion MNIST Label Summary}
\renewcommand{\arraystretch}{1.3} 
\centering
\begin{tabular}{|ccc|ccc|}
\hline
Label & Name & \# Instances& Label & Name & \# Instances\\ 
\hhline{|===|===|}
0 & T-shirt/top & 7000 & 5 & Sandal & 7000\\
1 & Trouser& 7000 & 6 & Shirt & 7000\\
2 & Pullover & 7000 & 7 & Sneaker & 7000\\
3 & Dress & 7000 & 8 & Bag & 7000\\
4 & Coat & 7000 & 9 & Ankle boot & 7000\\  
\hline
\end{tabular}
	\label{Tab:}
\end{table}
\item Attributes:  \textbf{784 attributes (pixels)}
\end{itemize}

\subsection{Dogs vs. Cats}
\noindent 
Dogs vs. Cats \cite{Kaggle} dataset released in 2013 is a dataset for a machine learning competition on Kaggle. The task of this dataset is to classify whether an animal in the image is a dog or a cat. There are 25,000 unequal-sized color images associated with labels from 2 classes. 

Since it requires a multivariate dataset serving as the input of the classification models considered, we first resize all images into $224 \times 224$ then extract the features via VGG16. After the extraction, we flatten the feature's shape of $7\times 7\times 512$ into $p = 25,088$. In the end, we obtain the Dogs vs. Cats dataset of shape $25,000\times 25,088$.

\begin{figure}[htbp]
\centering
\includegraphics[scale=0.35]{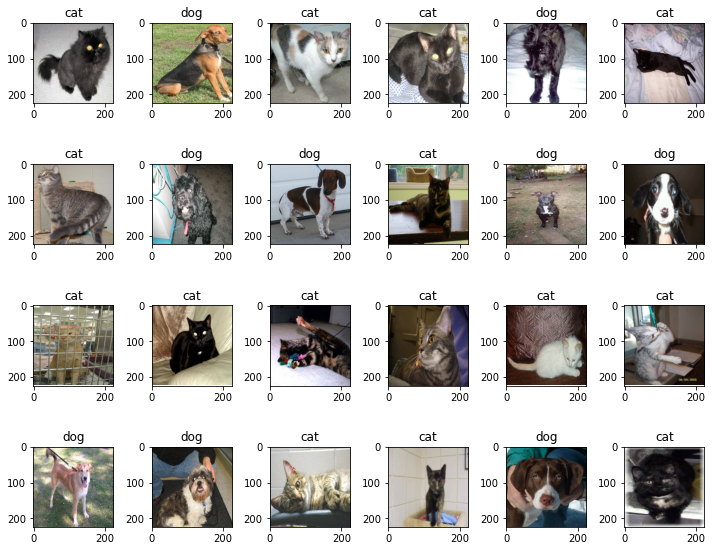}
\caption{Dogs vs. Cats Sample Data}
\label{Fig:}
\end{figure}

\begin{itemize}
\item Labels: \textbf{2 labels}
\begin{table}[htbp]
\caption{Dogs vs. Cats Label Summary}
\renewcommand{\arraystretch}{1.3} 
\centering
\begin{tabular}{|ccc|ccc|}
\hline
Label & Name & \# Instances & Label & Name & \# Instances\\ 
\hhline{|===|===|}
0 & Cat & 12,500& 1 & Dog & 12,500\\
\hline
\end{tabular}
	\label{Tab:}
\end{table}
\item Attributes:  \textbf{25,088 attributes (pixels)}
\end{itemize}

\subsection{Census Income}
\noindent 
Census Income dataset\cite{Dua:2019} also known as Adult dataset was donated to UCI Machine Learning Repository in 1996. The task of this dataset is to predict whether income exceeds 50k/yr based on census data. There are 48,842 examples with 14 attributes, associated with labels from 2 classes. After removing rows containing unknown values, there are 45,222 examples left.
\begin{itemize}
\item Labels: \textbf{2 labels}
\begin{table}[htbp]
\caption{Census Income Label Summary}
\renewcommand{\arraystretch}{1.3} 
\centering
\begin{tabular}{|ccc|ccc|}
\hline
Label & Name & \# Instances & Label & Name & \# Instances\\ 
\hhline{|===|===|}
0 & Income $\leq$ 50k & 34,014 & 1 & Income $>$ 50k & 11,208\\
\hline
\end{tabular}
	\label{Tab:}
\end{table}
\item Attributes: \textbf{14 attributes}
\begin{enumerate}[label={\arabic*)}]
\item \textit{age}: continuous.
\item \textit{workclass}: Private, Self-emp-not-inc, Self-emp-inc, Federal-gov, Local-gov, State-gov, Without-pay, Never-worked.
\item \textit{fnlwgt}: continuous.
\item \textit{education}: Bachelors, Some-college, 11th, HS-grad, Prof-school, Assoc-acdm, Assoc-voc, 9th, 7th-8th, 12th, Masters, 1st-4th, 10th, Doctorate, 5th-6th, Preschool.
\item \textit{education-num}: continuous.
\item\textit{ marital-status}: Married-civ-spouse, Divorced, Never-married, Separated, Widowed, Married-spouse-absent, Married-AF-spouse.
\item \textit{occupation}: Tech-support, Craft-repair, Other-service, Sales, Exec-managerial, Prof-specialty, Handlers-cleaners, Machine-op-inspct, Adm-clerical, Farming-fishing, Transport-moving, Priv-house-serv, Protective-serv, Armed-Forces.
\item \textit{relationship}: Wife, Own-child, Husband, Not-in-family, Other-relative, Unmarried.
\item \textit{race}: White, Asian-Pac-Islander, Amer-Indian-Eskimo, Other, Black.
\item \textit{sex}: Female, Male.
\item \textit{capital-gain}: continuous.
\item \textit{capital-loss}: continuous.
\item \textit{hours-per-week}: continuous.
\item \textit{native-country}: United-States, Cambodia, England, Puerto-Rico, Canada, Germany, Outlying-US(Guam-USVI-etc), India, Japan, Greece, South, China, Cuba, Iran, Honduras, Philippines, Italy, Poland, Jamaica, Vietnam, Mexico, Portugal, Ireland, France, Dominican-Republic, Laos, Ecuador, Taiwan, Haiti, Columbia, Hungary, Guatemala, Nicaragua, Scotland, Thailand, Yugoslavia, El-Salvador, Trinadad and Tobago, Peru, Hong, Holand-Netherlands.
\end{enumerate}
\end{itemize}

\subsection{Dry Bean}
\noindent 
Dry Bean dataset\cite{Koklu2020} was donated to UCI Machine Learning Repository in 2020. The dataset is derived from 13,611 images of 7 different dry beans. The 16 features in the images are extracted by a computer vision system including the form, shape, etc. The task of this dataset is to classify dry beans into correct classes. There are 13,611 examples with 16 attributes, associated with labels from 7 classes. 
\begin{itemize}
\item Labels: \textbf{7 labels}
\begin{table}[htbp]
\caption{Dry Bean Label Summary}
\renewcommand{\arraystretch}{1.3} 
\centering
\begin{tabular}{|ccc|ccc|}
\hline
Label & Name & \# Instances & Label & Name & \# Instances\\ 
\hhline{|===|===|}
0 & BARBUNYA & 1,322 & 4 & HOROZ & 1,928\\
1 & BOMBAY & 522 & 5 & SEKER & 2,027\\
2 & CALI & 1,630 & 6 & SIRA & 2,636\\
3 & DERMASON & 3,546 &  & & \\
\hline
\end{tabular}
	\label{Tab:}
\end{table}
\item Attributes: \textbf{16 attributes}
\begin{enumerate}[label={\arabic*)}]
\item \textit{Area (A)}: The area of a bean zone and the number of pixels within its boundaries.
\item \textit{Perimeter (P)}: Bean circumference is defined as the length of its border.
\item \textit{Major axis length (L)}: The distance between the ends of the longest line that can be drawn from a bean.
\item \textit{Minor axis length (l)}: The longest line that can be drawn from the bean while standing perpendicular to the main axis.
\item \textit{Aspect ratio (K)}: Defines the relationship between L and l.
\item \textit{Eccentricity (Ec)}: Eccentricity of the ellipse having the same moments as the region.
\item \textit{Convex area (C)}: Number of pixels in the smallest convex polygon that can contain the area of a bean seed.
\item \textit{Equivalent diameter (Ed)}: The diameter of a circle having the same area as a bean seed area.
\item \textit{Extent (Ex)}: The ratio of the pixels in the bounding box to the bean area.
\item \textit{Solidity (S)}: Also known as convexity. The ratio of the pixels in the convex shell to those found in beans.
\item \textit{Roundness (R)}: Calculated with the following formula: $(4\pi A)/(P^2)$
\item \textit{Compactness (CO)}: Measures the roundness of an object: $Ed/L$
\item \textit{ShapeFactor1 (SF1)}
\item \textit{ShapeFactor2 (SF2)}
\item \textit{ShapeFactor3 (SF3)}
\item \textit{ShapeFactor4 (SF4)}
\end{enumerate}
\end{itemize}

\section{Performance Evaluation}
\noindent 
Each datasets is divided into training dataset, validation dataset, and testing dataset. We consider 10-fold cross validation for model tuning. In addition, accuracy is set to be the evaluation metric.

\begin{table}[htbp]
\caption{Dataset Splitting Summary} 
\renewcommand{\arraystretch}{1.4} 
\centering
\begin{tabular}{cccccc} 
\toprule
Name & \# Instances & Training Dataset & Validation Dataset & Testing Dataset\\ \midrule
MNIST & 70,000 & 54,000 & 6,000& 10,000\\
Fashion MNIST & 70,000 & 54,000 & 6,000& 10,000\\ 
Dogs vs. Cats & 25,000 & 15,750 & 1,750 & 7,500\\
Census Income & 45,222 & 27,146 & 3,016 & 15,060\\ 
Dry Bean & 13,611 & 8,574  & 953 & 4,084\\ 
\bottomrule
\end{tabular}
	\label{Tab:}
\end{table}

The following tables present the average and standard deviation of 10-fold cross validation's accuracies among QMS and 6 candidate algorithms over 5 different datasets. In addition, we also report the testing accuracy. The testing accuracy is calculated from the model constructed by whole training+validation dataset deploying on the testing dataset. Validation dataset in each fold is used to tune the hyperparameters aiming to achieve better performance in testing dataset. For those datasets (Dogs vs. Cats, Dry Bean) which are not divided originally from the sources, we split them by 7:3 for training+validation and testing. Note that in the following tables, we color the top 3 classification methods in blue for each dataset.

\clearpage

\begin{table}[htbp]
\caption{Average of 10-Fold  CV's Accuracy}
\renewcommand{\arraystretch}{1.2} 
\centering
\begin{tabular}{cccccccc} 
\toprule
 &QMS & k-NN &Logistic &SVM& RF& XGB & ANN \\ \midrule
MNIST   & 96.98\%  & 94.71\% & 91.45\%& 96.48\% & {\color{blue}97.06\%} & {\color{blue}98.03\%} & {\color{blue}97.06\%} \\
Fashion MNIST & 88.60\% & 85.66\% & 84.68\%& {\color{blue}89.30\%}  &88.59\%&  {\color{blue}91.17\%}& {\color{blue}89.15\%}\\ 
Dogs vs. Cats & {\color{blue}92.65\%}  & 72.62\% & 92.36\% & {\color{blue}92.75\%} & 89.90\%  & 92.48\% & {\color{blue}92.58\%}\\ 
Census Income & 84.97\% & 83.63\%  & 84.83\%  & {\color{blue}85.39\%} & {\color{blue}85.96\%} & {\color{blue}86.98\%}  & 85.07\%\\ 
Dry Bean  & {\color{blue}92.98\%} & 92.61\%  & {\color{blue}92.65\%}  & {\color{blue}93.19\%} & 92.21\% & 92.60\% & 92.47\%  \\ \bottomrule
\end{tabular}
	\label{Tab:val_avg}
\end{table}

\begin{table}[htbp]
\caption{Standard Deviation of 10-Fold  CV's Accuracy}
\renewcommand{\arraystretch}{1.2} 
\centering
\begin{tabular}{cccccccc}
\toprule
 & QMS & k-NN & Logistic & SVM & RF & XGB & ANN \\ \midrule
MNIST  & \llapph{0.17\%} & \llapph{0.31\%} & \llapph{0.27\%} & \llapph{0.23\%} & \llapph{0.15\%} &  \llapph{0.14\%} & \llapph{0.19\%} \\
Fashion MNIST & \llapph{0.59\%} & \llapph{0.27\%} & \llapph{0.33\%} & \llapph{0.16\%}  &\llapph{0.30\%} & \llapph{0.27\%} & \llapph{0.37\%}\\ 
Dogs vs. Cats &  \llapph{0.69\%} &  \llapph{1.02\%} &  \llapph{0.57\%} & \llapph{0.71\%} & \llapph{0.27\%} & \llapph{0.48\%} &  \llapph{0.52\%}\\ 
Census Income & \llapph{0.45\%} & \llapph{0.74\%}  & \llapph{0.60\%}  & \llapph{0.75\%} & \llapph{0.70\%} & \llapph{0.65\%}  & \llapph{0.42\%}\\ 
Dry Bean  & \llapph{0.47\%} & \llapph{0.54\%}  & \llapph{0.91\%}  & \llapph{0.88\%} & \llapph{0.76\%} & \llapph{0.85\%} & \llapph{0.71\%}\\ \bottomrule
\end{tabular}
	\label{Tab:val_sd}
\end{table}

\begin{table}[htbp]
\caption{Testing Accuracy}
\renewcommand{\arraystretch}{1.2} 
\centering
\begin{tabular}{cccccccc}
\toprule
 & QMS & k-NN & Logistic & SVM & RF & XGB & ANN \\ \midrule
MNIST  & {\color{blue}97.33\%}  & 94.43\% & 92.12\%& 96.60\% & 97.20\% & {\color{blue}98.11\%} & {\color{blue}97.23\%} \\
Fashion MNIST & {\color{blue}88.44\%} & 85.33\% & 83.44\%& 88.36\%  &88.05\%& {\color{blue}90.63\%} & {\color{blue}88.76\%}\\ 
Dogs vs. Cats &  {\color{blue}93.77\%} & 73.33\% & 92.69\% & {\color{blue}93.07\%}  &89.97\%  &  93.05\%& {\color{blue}93.16\%}\\ 
Census Income & 84.69\% & 83.76\%  & 83.84\%  & {\color{blue}85.19\%} & {\color{blue}85.85\%} & {\color{blue}87.07\%}  & 84.76\%\\ 
Dry Bean  & {\color{blue}92.63\%} & {\color{blue}92.43\%}  & 91.99\%  & 92.24\% & 91.43\% & {\color{blue}92.61\%} & 92.38\%\\ \bottomrule
\end{tabular}
	\label{Tab:test}
\end{table}

From Table \ref{Tab:val_avg}, we can see that gradient-based QMS performs the best on Dry Bean dataset. In addition, it outperforms k-NN and logistic regression on all datasets. As for testing accuracy shown in Table \ref{Tab:test}, gradient-based QMS is the winner on Dry Bean and Dogs vs. Cats dataset, and places top 3 on MNIST and Fashion MNIST datasets. Table \ref{Tab:val_sd} presents that the gradient-based QMS's standard deviation of 10-fold cross validation's accuracies is comparable to the rest since there is no abnormal value detected. Note that the number of parameters used in ANN is set to be closed to that in QMS.   

\clearpage
\chapter{Conclusion}

\noindent 
In the thesis, we firstly introduce the structure of Quadratic Multiform Separation (QMS), and then incorporate the gradient-based optimization algorithm Adam to obtain a classifier for the underlining classification problem. Moreover, we analyze the relationship between hyperparameter and accuracy, and provide recommendations regarding model tuning. Lastly, we compare the empirical performance with six matured classification methods on five popular datasets. 

The gradient of the QMS-specific loss function is derived using the techniques of matrix algebra. The application of Adam allows us to deal with the complex error surface of QMS-specific loss function. As for model tuning, we conclude that there is a slightly curvature pattern of accuracy with respect to control hyperparameter $\alpha$, the peak tends to present at the point that the $\alpha$ is not more than $0.5$, moreover, the accuracy diminishes dramatically after the correponding $\alpha$ of the peak; in regard to weight hyperparameter $q$, we conclude that the accuracy significantly increases while $q$ is not larger than $5$, but one can still pursue for slightly higher accuracy with greater $q$. Finally, the empirical results show that the performance of our gradient-based QMS is comparable to commonly-used classification methods, since our performance is pretty close to these methods' and even steps into the first three places under some of the classification problems in terms of accuracy.

\clearpage
\bibliographystyle{unsrt}
\bibliography{Reference}

\begin{thebibliography}{10}

\bibitem{1403109}
Phil Simon.
\newblock {\em Too Big to Ignore: The Business Case for Big Data}.
\newblock Wiley, 1 edition, 2013.

\bibitem{27308}
Tom~M. Mitchell.
\newblock {\em Machine Learning}.
\newblock McGraw-Hill series in computer science. McGraw-Hill, 1 edition, 1997.

\bibitem{FanJanuary142021}
Ko-Hui~Michael Fan, Chih-Chung Chang, Ye-Hwa Chen, and Kuang-Hsiao-Yin
  Kongguoluo, US. Patent 17/148,860, filed January 14, 2021.

\bibitem{FanApril142021}
Ko-Hui~Michael Fan, Chih-Chung Chang, and Kuang-Hsiao-Yin Kongguoluo, US.
  Patent 17/230,283, filed April 14, 2021.

\bibitem{917808}
Gareth James, Daniela Witten, Trevor Hastie, and Robert Tibshirani.
\newblock {\em An Introduction to Statistical Learning with Applications in R}.
\newblock Springer Texts in Statistics, Vol. 103. Springer, 2013.

\bibitem{Breiman2001}
Leo Breiman.
\newblock Random forests.
\newblock {\em Machine Learning}, 45(1):5--32, 2001.

\bibitem{Raileanu2004}
Laura Raileanu and Kilian Stoffel.
\newblock Theoretical comparison between the gini index and information gain
  criteria.
\newblock {\em Annals of Mathematics and Artificial Intelligence}, 41:77--93,
  05 2004.

\bibitem{2158511}
Galit Shmueli, Peter~C. Bruce, Inbal Yahav, Nitin~R. Patel, and Kenneth
  C.~Lichtendahl Jr.
\newblock {\em Data Mining for Business Analytics: Concepts, Techniques, and
  Applications in R}.
\newblock Wiley, 1 edition, 2017.

\bibitem{Noble2006}
William~S Noble.
\newblock What is a support vector machine?
\newblock {\em Nature biotechnology}, 24(12):1565—1567, December 2006.

\bibitem{DBLP:journals/corr/ChenG16}
Tianqi Chen and Carlos Guestrin.
\newblock Xgboost: {A} scalable tree boosting system.
\newblock {\em CoRR}, abs/1603.02754, 2016.

\bibitem{Jain}
Ani1~K. Jain, Jianchang Mao, and K.M. Mohiuddin.
\newblock Artificial neural networks: A tutorial.
\newblock {\em IEEE}, 1996.

\bibitem{kingma2017adam}
Diederik~P. Kingma and Jimmy Ba.
\newblock Adam: A method for stochastic optimization, 2017.

\bibitem{lecun-mnisthandwrittendigit-2010}
Yann LeCun and Corinna Cortes.
\newblock {MNIST} handwritten digit database.
\newblock 2010.

\bibitem{xiao2017fashionmnist}
Han Xiao, Kashif Rasul, and Roland Vollgraf.
\newblock Fashion-mnist: a novel image dataset for benchmarking machine
  learning algorithms, 2017.

\bibitem{Kaggle}
Kaggle.
\newblock Dogs vs. cats.
\newblock \url{https://www.kaggle.com/c/dogs-vs-cats/overview}, 2013.

\bibitem{Dua:2019}
Dheeru Dua and Casey Graff.
\newblock {UCI} machine learning repository - census income, 2017.

\bibitem{Koklu2020}
Murat Koklu and Ilker~Ali Ozkan.
\newblock Multiclass classification of dry beans using computer vision and
  machine learning techniques.
\newblock {\em Computers and Electronics in Agriculture}, 174:105507, 2020.

\end{thebibliography}

\end{document}